\documentclass{article} %
\usepackage[preprint]{arxiv_style}

\usepackage{microtype}
\usepackage{hyperref}
\usepackage{url}
\usepackage{booktabs}
\usepackage{graphicx}
\usepackage{lineno}
\usepackage{amsmath}
\usepackage{amsmath}
\usepackage{xcolor}
\usepackage{tcolorbox}
\usepackage{multirow}
\usepackage{xspace}

\usepackage{listings}
\usepackage{xcolor}

\usepackage{wrapfig}

\definecolor{codegreen}{rgb}{0,0.6,0}
\definecolor{codegray}{rgb}{0.5,0.5,0.5}
\definecolor{codepurple}{rgb}{0.58,0,0.82}
\definecolor{backcolour}{rgb}{0.95,0.95,0.92}

\lstdefinestyle{mystyle}{
    backgroundcolor=\color{backcolour},
    commentstyle=\color{codegreen},
    keywordstyle=\color{magenta},
    stringstyle=\color{codepurple},
    basicstyle=\ttfamily\small,
    breakatwhitespace=false,
    breaklines=true,
    captionpos=b,
    keepspaces=true,
    numbersep=5pt,
    showspaces=false,
    showstringspaces=false,
    showtabs=false,
    tabsize=2,
    frame=single,
    rulecolor=\color{codegray},
    framesep=3pt,
    framerule=0.4pt,
}

\newtcolorbox{promptbox}{
    colback=blue!5,    %
    colframe=blue!50,  %
    boxrule=0.8pt,     %
    arc=4pt,           %
    left=5pt,          %
    right=5pt,         %
    top=5pt,           %
    bottom=5pt,        %
    width=\linewidth   %
}

\definecolor{darkblue}{rgb}{0, 0, 0.5}
\hypersetup{colorlinks=true, citecolor=darkblue, linkcolor=darkblue, urlcolor=darkblue}

\lstdefinestyle{pythontitle}{
    language=Python,
    basicstyle=\% \large\ttfamily,  %
    keywordstyle=\color{blue},
    commentstyle=\color{green!60!black},
    stringstyle=\color{orange},
    showstringspaces=false,
}

\title{Sleep-time Compute: Beyond Inference Scaling at Test-time}

\author{\textbf{Kevin Lin} \textsuperscript{1}$^* $
\hspace{2pt} \textbf{Charlie Snell} \textsuperscript{2}$^*$ \\
\textbf{Yu Wang} \textsuperscript{1}
\quad
\textbf{Charles Packer} \textsuperscript{1}
\quad
\textbf{Sarah Wooders} \textsuperscript{1}
\quad
\textbf{Ion Stoica} \textsuperscript{1} \textsuperscript{2}
\quad
\textbf{Joseph E. Gonzalez} \textsuperscript{1} \textsuperscript{2}\\[5pt]
\textsuperscript{1}Letta \quad \textsuperscript{2}University of California, Berkeley \\[3pt]
\texttt{research@letta.com} \\
}

\newcommand{\name}{sleep-time compute\xspace}

\newcommand{\gsmkdataset}{Stateful GSM-Symbolic}
\newcommand{\aimedataset}{Stateful AIME}
\newcommand{\gsmkamortizationdataset}{Multi-Query GSM-Symbolic}
\newcommand{\codedataset}{SWE-Features}

\newif\ifcomments
\commentstrue
\ifcomments
    \newcommand{\kevin}[1]{{\bf\color{blue} [kevin: #1]}}
    \newcommand{\ion}[1]{{\bf\color{blue} [ion: #1]}}
    \newcommand{\joey}[1]{{\bf\color{orange} [joey: #1]}}

\else
    \newcommand{\kevin}[1] {}
    \newcommand{\ion}[1] {}
    \newcommand{\joey}[1] {}
\fi

\begin{document}

\maketitle

\begin{abstract}

Scaling test-time compute has emerged as a key ingredient for enabling large language models (LLMs) to solve difficult problems, but comes with high latency and inference cost. 
We introduce \name{}, which allows models to ``think'' offline about contexts before queries are presented: by anticipating what queries users might ask and pre-computing useful quantities, we can significantly reduce the compute requirements at test-time. To demonstrate the efficacy of our method, we create modified versions of two reasoning tasks -- \gsmkdataset{} and \aimedataset{}. We find that \name{} can reduce the amount of test-time compute needed to achieve the same accuracy by $\sim 5\times$ on \gsmkdataset{} and \aimedataset{} and that by scaling \name{} we can further increase accuracy by up to 13\% on \gsmkdataset{} and 18\% on \aimedataset{}. Furthermore, we introduce \gsmkamortizationdataset{}, which extends GSM-Symbolic by including multiple related queries per context. By amortizing \name{} across related queries about the same context using \gsmkamortizationdataset{}, we can decrease the average cost per query by $2.5\times$. We then conduct additional analysis to understand when \name{} is most effective, finding the predictability of the user query to be well correlated with the efficacy of \name{}. Finally, we conduct a case-study of applying \name{} to a realistic agentic SWE task. Code and data released at: \href{https://github.com/letta-ai/sleep-time-compute}{https://github.com/letta-ai/sleep-time-compute}.

\end{abstract}

\section{Introduction}

\begin{figure}[ht]
    \centering
        \includegraphics[width=1.01\textwidth]{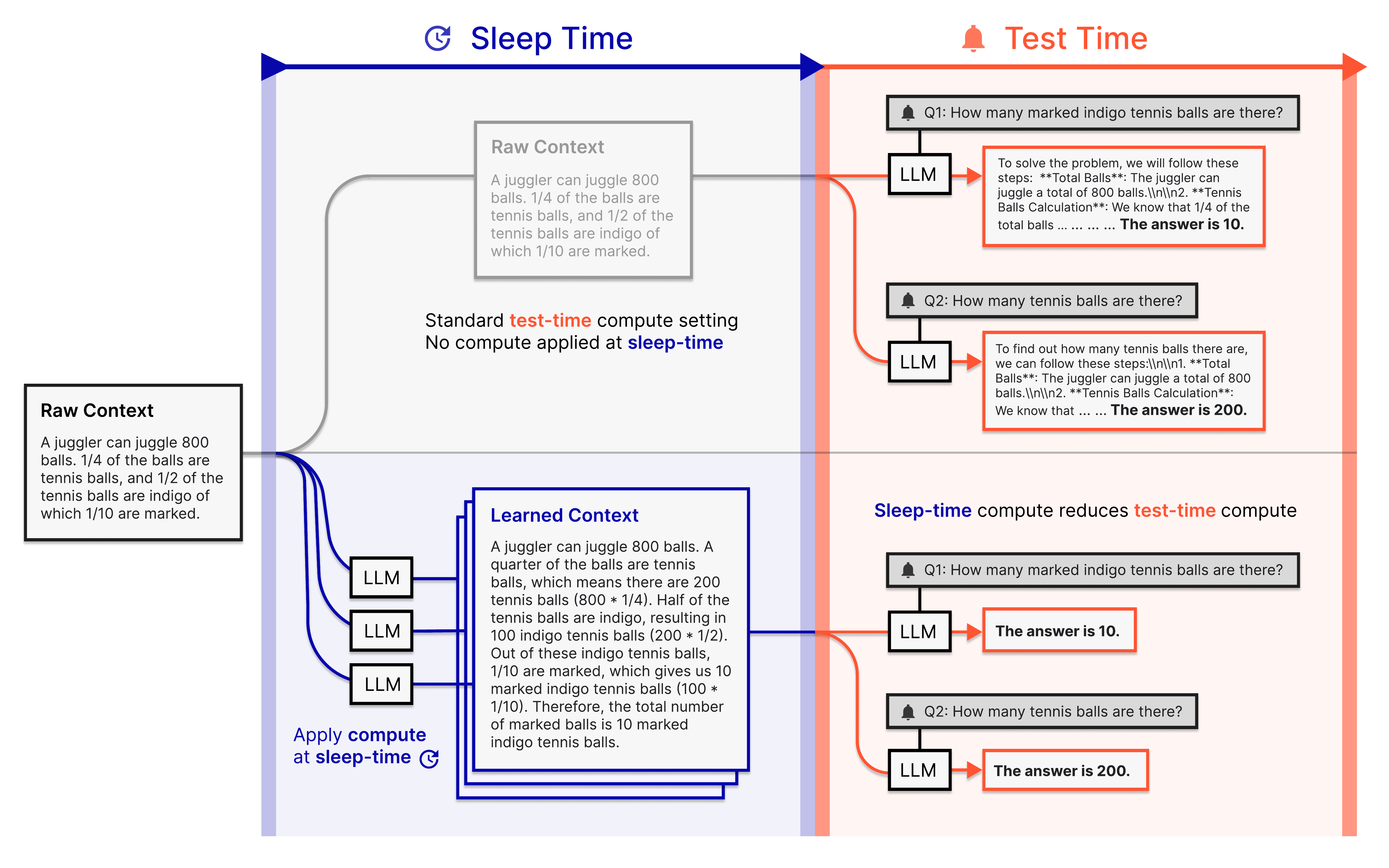}
     
    \caption{Example of applying \name{} on \gsmkamortizationdataset{}-P1. Sleep-time compute processes the original raw context, adding additional computations that can potentially be useful for future queries. Moreover, contexts can be shared across related queries enabling savings in total cost per query.}
    \label{fig:box}
\end{figure}

Test-time scaling has emerged as an effective way to boost LLM performance on challenging tasks by spending more time thinking on difficult problems \citep{openai2024openaio1card,deepseek2024r1,snell2024scalingllmtesttimecompute, brown2024large}. 
However, improved performance from test-time compute comes at a significant increase in latency and cost, waiting potentially several minutes for answers and costing up to tens of dollars per query.\footnote{https://platform.openai.com/docs/models/o1-pro} These drawbacks are in part due to the fact that the current approach to applying test-time compute assumes that problems are stateless, i.e. queries (user queries at test-time) and the contexts (background information) required for answering them are provided to the model together at ``test-time.'' In practice, this means that if multiple related queries require making similar inferences about the context at ``test-time,'' the model will have to recompute redundant computations each time, incurring additional latency and cost.

In reality, many LLM applications are \emph{inherently stateful}, and work in conjunction with persisted, re-used context. A classic example is document question-answering, where documents contextualize responses to questions. Coding agents also operate on a large common repository and participate in multiple rounds of debugging support, while conversational assistants need to maintain the past dialogue. In all these applications, there is context (available documents, a codebase, or conversation history) that is already available before the next user input.

In these settings, we could in principle, make useful inferences about the current state (context) offline before, or even during the user's next input. We refer to such a process, as \name{}: where inference is done between interactions with the model while it would otherwise be idle in \textit{sleep}-time. In practice, this is achieved by prompting the model to generate a new context consisting of inferences about the existing context, which may be potentially useful for answering test-time queries. The re-represented context from sleep-time can then be provided in the prompt at test-time, enabling the model to respond to user queries at the accuracy of standard test-time compute but with far lower latencies. 
For example, a coding assistant at sleep-time may identify architectural patterns, anticipate potential debugging strategies, or infer optimizations prior to the user input. Moreover, users might ask multiple queries about the same context. In these settings, any inferences made during sleep-time can be shared across queries, effectively amortizing the cost of sleep-time compute and reducing the total average cost per query.

To evaluate \name{}, we modify two mathematical reasoning datasets to introduce two datasets -- \gsmkdataset{} and \aimedataset{} -- by splitting the existing problems in these datasets into a context and a question. Using these datasets, we aim to empirically understand the benefits of \name{} on standard test-time compute benchmarks. We show that:
\begin{itemize}
    \setlength{\itemsep}{0pt}     %
    \setlength{\parskip}{0pt}     %
    \setlength{\parsep}{0pt}      %
    \setlength{\leftmargin}{1pt} %
    \item Sleep-time compute produces a pareto improvement in the test-time compute vs. accuracy curve, reducing the test-time compute needed to achieve the same accuracy by $\sim 5\times$ on \gsmkdataset{} and \aimedataset{}.
    \item By scaling up \name{}, we see further pareto improvements, shifting the accuracy up by 13\% on \gsmkdataset{} and 18\% on \aimedataset{}.
    \item By amortizing \name{} across multiple queries for the same context, we can reduce the average cost per question by $2.5\times$.
    \item We conduct analysis to understand which queries benefit the most from \name{}, finding that \name{} is more effective in settings where the query is more easily predictable from the context.
\end{itemize}
Finally, we end with case study of applying \name{} to reduce test-time compute in a realistic agentic software engineering task.

\section{Related Work}
\paragraph{Scaling test-time compute.} 
Our work builds on recent progress on scaling up computation at test-time for difficult reasoning problems \citep{snell2024scalingllmtesttimecompute,deepseek2024r1,openai2024openaio1card}. Two predominant approaches to test-time scaling have emerged: sequential test-time scaling~\citep{openai2024openaio1card,deepseek2024r1,muennighoff2025s1simpletesttimescaling,snell2024scalingllmtesttimecompute} and parallel test-time scaling~\citep{brown2024large,snell2024scalingllmtesttimecompute}. While sequential test-time scaling has demonstrated impressive performance improvements, parallel test-time scaling has the advantage of scaling test-time compute without increasing latency. In constrast, we propose an alternative dimension where existing advancements in test-time compute, both sequential and parallel can be applied. Namely, instead of performing inference purely at test-time, we leverage compute on contexts that are available before the actual query arrives.

\paragraph{Speculative decoding in LLMs.} Speculative decoding is a standard technique for reducing latency in decoding with LLMs ~\citep{leviathan2023fastinferencetransformersspeculative,stern2018blockwiseparalleldecodingdeep,cai2024medusasimplellminference,deepseekai2025deepseekv3technicalreport}. Sleep-time compute similarly targets reducing reasoning latency by speculating on the \emph{user's query} as well as any potentially helpful reasoning over the context. However, unlike speculative decoding, the generated tokens are used as an input regardless of the user's actual query, and at test-time the reasoning model uses these generated tokens to help answer the user query more efficiently.

\paragraph{Pre-computation.} 
Beyond LLMs, a long history of work has explored the trade-off between pre-computation and memory (eg. memory caches \cite{smith1982cache} and data cubes for OLAP workloads \cite{gray1997data}). Our work explores the same trade-off between query latency and pre-computation overhead, operating under the assumption that query workload patterns can be reasonably anticipated in advance. \name{} builds on the idea of pre-fetching in traditional operating systems, in the context of LLMs à la \cite{packer2023memgpt}, storing frequently used computational results to avoid higher latency at test-time.

\section{Sleep-time Compute}

In the standard paradigm of applying test-time compute, a user inputs a prompt $p$ to the LLM and then the LLM applies test-time compute to help answer the user's question. However, the $p$ provided to the LLM can oftentimes be decomposed into a pre-existing context $c$ (eg. a codebase) and a user query $q$ (eg. a question about the codebase). When the LLM is not actively responding to the user, it typically still has access to the existing context $c$. During this time, the LLM is typically idling, missing the opportunity to reason about $c$ offline: a process we term \name{}.

\paragraph{Test-time compute.} In the test-time compute setting, the user provides $q$ along with some context $c$ and the model outputs a reasoning trace followed by a final answer $a$. We denote this process, as: $T_{B}(q, c) \rightarrow a$,
where $T$ is the method for using test-time compute with budget $B$, which could include techniques like extended chains of thought or best-of-N. In practice, the user may have multiple queries about the same context $q_1$, $q_2$ ... $q_N$. In this setting, the model will carry out independent reasoning processes for each $q_i$, even if they are related to the same context $c$. Ideally, we would be able to reuse related inferences across each $q_i$ to save compute. Moreover, in many cases, $c$ is complex and may require carrying out significant processing/inferences in order to provide an answer to $q$. 
Since, the test-time compute paradigm of $T(q, c) \rightarrow a$ assumes that $c$ is only available at the same time as $q$, standard test-time compute carries out all of these inferences only after the user provides the query, causing the user to wait up to several minutes for a response. 
However, in practice we often have access to $c$ before $q$ and can carry out much of this processing ahead of time.

\paragraph{Sleep-time compute.} During sleep-time we are given the context $c$ but not the query $q$. 
Using just this context $c$, we can use the LLM to 
infer likely questions and reason about the context ultimately producing a more new re-represented context $c'$.
We denote this process as: $S(c) \rightarrow c'$, where $S$ can be any standard test-time scaling technique applied towards pre-processing the context at sleep-time. In this work, $S(c)$ is implemented by prompting the model to draw inferences and re-write $c$ in a way that might be useful at test-time (see Appendix~\ref{app:implementation_details} for more details). After pre-processing the context, we can provide the new context $c'$ at test-time in place of $c$ to produce a final answer to the user's query: $T_{b}(q, c') \rightarrow a$. Since much of the reasoning about $c$ has been done ahead of time in this case, we can use a much smaller test-time budget $b << B$. Moreover, $c'$ can be shared across different queries $q_i$ about the same context, effectively amortizing the compute required to arrive at $c'$ across queries, providing a total cost saving.

\section{Experimental Setup}

Next, we describe the datasets, models, and baselines we use to evaluate \name{}.

\subsection{Datasets}

We select datasets which represent standard benchmarks for LLM reasoning and test-time scaling, and which demonstrate improvements from scaling test-time compute with state-of-the-art LLMs (either reasoning or non-reasoning).

\begin{figure}

    \centering
    \includegraphics[width=\textwidth]{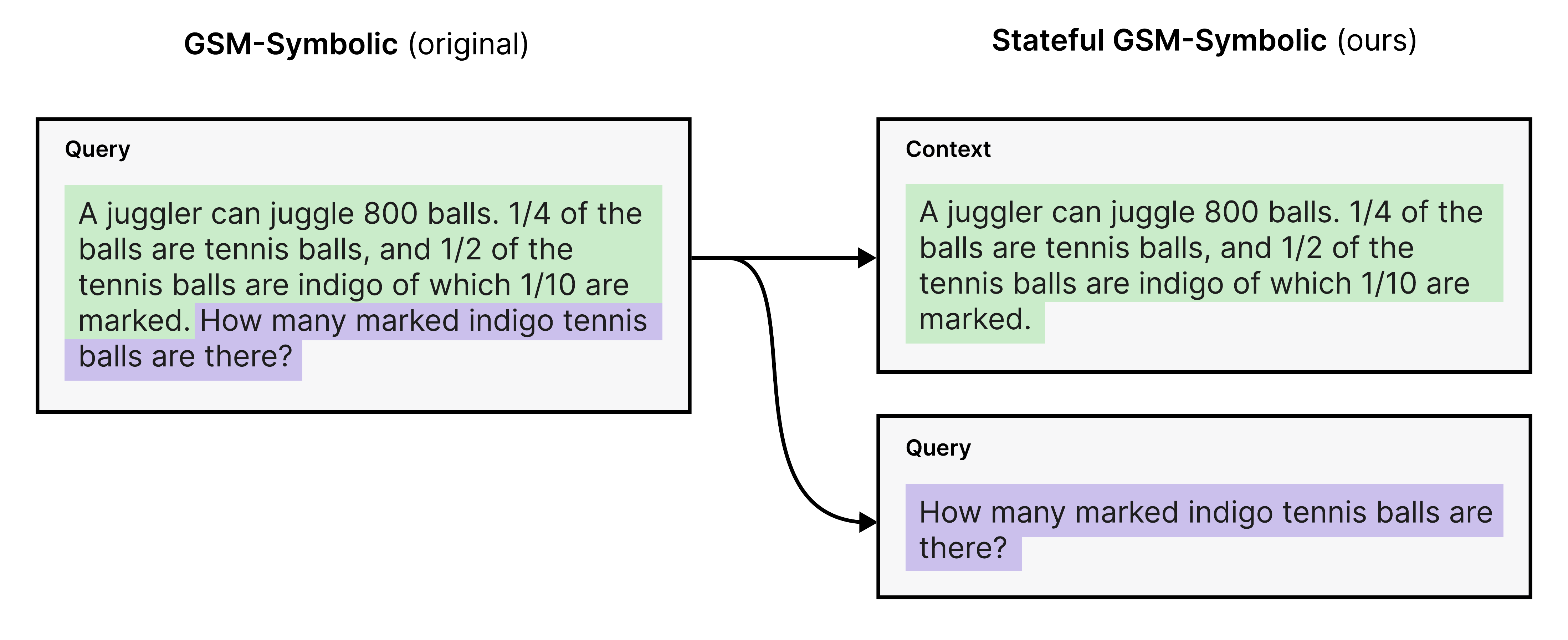}     
    \caption{Example of  separating an instance from GSM-Symbolic into context, and question, creating an instance in \gsmkdataset{}.}
    
    \label{fig:gsm8k_split}
\end{figure}

\paragraph{Stateful datasets.} We introduce two datasets to study applying \name{} in stateful settings, \gsmkdataset{}, and \aimedataset{}, where each dataset is derived from splitting the existing  datasets into a context and a question (see Figure~\ref{fig:gsm8k_split} for an example).
\gsmkdataset{} is derived from the P1 and P2 splits of  GSM-Symbolic  \citep{mirzadeh2024gsm}, which add one and two clauses respectively to the original GSM8K dataset \citep{cobbe2021training} to that increase the difficulty. GSM-Symbolic P1 contains 5000 examples and P2 2500 examples. \aimedataset{} contains 60 questions combined from AIME 2024 and 2025. In Appendix~\ref{app:aime_main_by_year} and~\ref{app:aime_scaling_by_year}, we show the breakdown of our results across AIME 2024 and 2025.

\paragraph{Amortization dataset.} To study the effect of related questions that share context, we introduce a new dataset \gsmkamortizationdataset{},  where each context has multiple queries. To generate multiple queries for a given context, we take \gsmkdataset{} and use o3-mini to generate additional question answer pairs. We synthetically generate additional questions from existing context question pairs in GSM-Symbolic. Appendix \ref{app:multi-query-gsm8k-symbolic} shows the prompt used to generate the additional questions. Figure~\ref{fig:multiquery_gsm8ksymbolic-examples} shows examples contexts and set of questions from the \gsmkamortizationdataset{} dataset and Table \ref{tab:multiquery_gsm8ksymbolic-stats} shows the overall dataset statistics.

\subsection{Models and Baselines}

\paragraph{Models.}  On each dataset, we evaluate models which have poor performance when using a small amount of test-time compute, but yield improvements from scaling up test-time compute. Therefore, on GSM-Symbolic, we conduct experiments using GPT-4o-mini and GPT-4o, and on AIME, we conduct experiments using OpenAI's o1, o3-mini, Anthropic's Claude Sonnet 3.7 Extended Thinking , and Deepseek-R1 \citep{deepseek2024r1}.~\footnote{https://openai.com/o1/} \footnote{https://www.anthropic.com/claude/sonnet}

\paragraph{Baselines} The main baseline we consider is the standard test-time compute setting in which both $c$ and $q$ are presented to the model for the first time at test-time. Furthermore, to validate that $q$ is not trivially predictable from $c$ on our \gsmkdataset{} and \aimedataset{} datasets, we also compare to a context-only baseline in Appendix~\ref{app:no_question_baseline}, in which the model is only given $c$ and is tasked with directly guessing an answer to the question it guesses is most likely to come next.

\section{Experiments and Results}

In this section, we carry out experiments to understand the benefits of \name{}. Specifically, we would like to answer each of the following questions using the math reasoning benchmarks introduced above:

\begin{enumerate}
    \setlength{\itemsep}{0pt}     %
    \setlength{\parskip}{0pt}     %
    \setlength{\parsep}{0pt}      %
    \setlength{\leftmargin}{1pt} %
    \item Can \name{} shift the pareto frontier of test-time compute vs. accuracy?
    \item Does scaling  \name in-turn improve the pareto further?
    \item When there are multiple related questions for a single context, can amortizing test-time compute with \name{} provide a total token efficiency benefit?
    \item In what settings does \name{} provide the most uplift?
\end{enumerate}

\subsection{Improving Pareto Test-Time Trade-off with \name}
\label{sec:sleep_pareto}

\begin{figure}
    \centering
    \includegraphics[width=\textwidth]{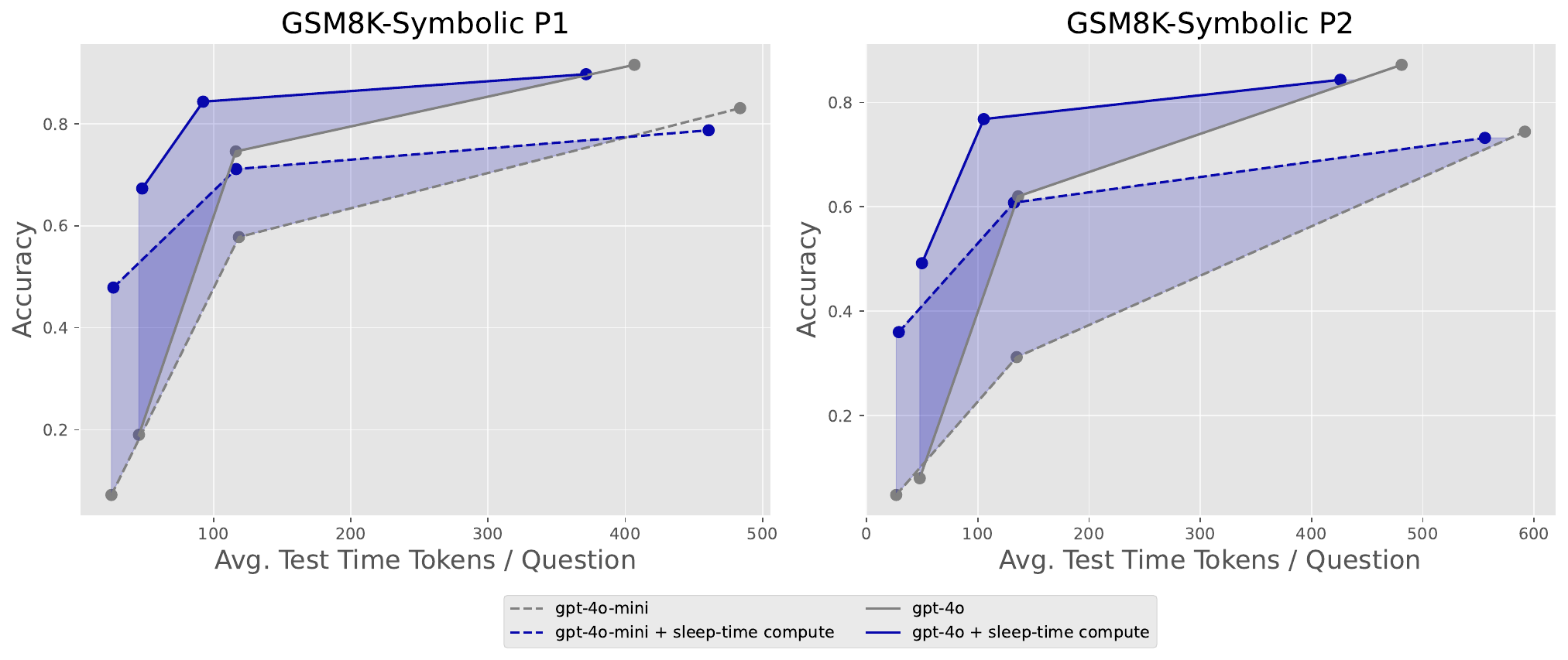} 
    \caption{The test-time compute vs. accuracy tradeoff for on \gsmkdataset{}. Shaded area indicates where \name improves the pareto test-time accuracy trade-off.}
    \label{fig:gsm8k-main-result}
\end{figure}

\begin{figure}[h!]
    \centering
    \includegraphics[width=\textwidth]{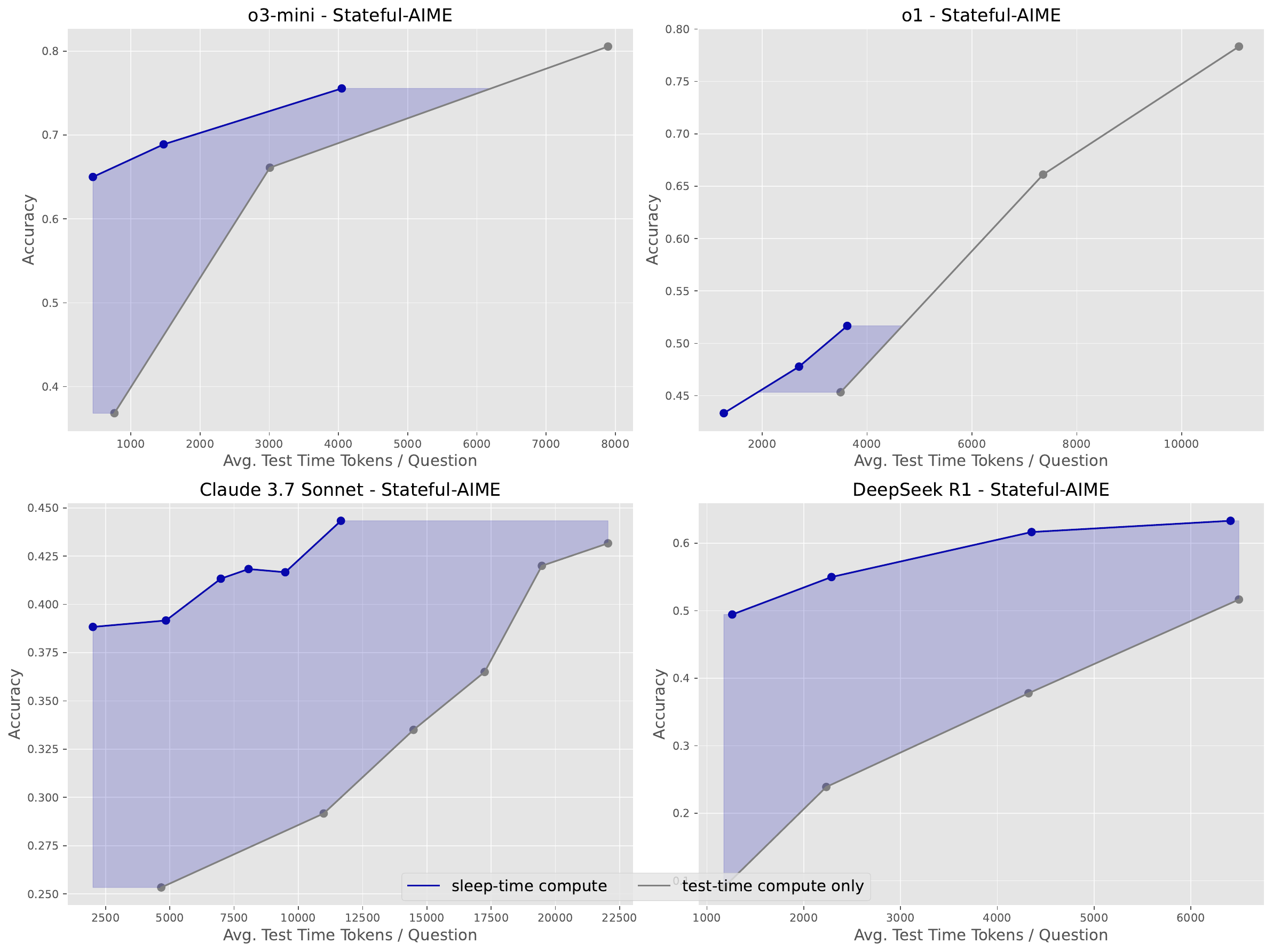}    
    \caption{The test-time compute vs. accuracy tradeoff on \aimedataset{} for various reasoning models. Applying \name{} allows models to reach similar levels of performance with much less compute at test-time. The shaded area indicates the pareto improvement from \name{}.}
    \label{fig:aime-main-results}
\end{figure}
We first determine the test-time compute, accuracy pareto  frontier by scaling standard test-time compute sequentially and in parallel. We then study how applying \name{} affects the pareto trade-off.

\paragraph{Scaling test-time-compute sequentially.} For non-reasoning models (GPT-4o and 4o-mini) on \gsmkdataset{}, to vary the amount of test-time compute, we construct prompts that instruct the model to use different amounts of verbosity at test time, eg. ``answer directly with a single sentence'' vs. ``double check your reasoning before outputting the final answer.'' The full prompts are in Appendix \ref{app:prompts}. We use temperature 0 for generation. We see in Figure \ref{fig:gsm8k-main-result} that there is a tradeoff between accuracy and the amount of test-time compute, and that adding \name{} can move beyond the pareto compute-accuracy curve. In particular, at lower test-time budgets, the performance of \name{} is significantly better than the baseline, achieving performance comparable to that of the baseline with $5\times$ less test-time tokens. However, at the test-tome compute budgets, the test-time compute only baseline slightly outperforms \name{}. We hypothesize that this may be because the standard test-time compute only has the content relevant to the specific question, so there is less distracting information in the prompt.

For reasoning models on \aimedataset{}, we scale the amount of test-time compute based on what is available in the API in the case of o1, o3-mini and Claude Sonnet 3.7. Since the Deepseek-R1 API does not provide a way to control  test-time compute, we apply the "budget forcing" and extension prompt from~\cite{muennighoff2025s1simpletesttimescaling}. Figure \ref{fig:aime-main-results} shows the results for each model on \aimedataset{}. We average results over 3 runs for o1, o3-mini and R1. For Claude 3.7 Sonnet, we average over 10 runs as we observed more noise in initial experiments. On all models, we see a significant test-time, accuracy pareto shift from applying \name{}, with the exception of o1, which demonstrates limited gains.
\paragraph{Scaling test-time compute in parallel.} An alternative approach to scaling test-time compute is via parallel sampling, which also has the benefit of maintaining low inference latency. The simplest approach to scaling parallel test-time compute is pass@k~\citep{brown2024large}, which makes the unrealistic assumption of having oracle query access to a ground truth verifier at test-time, an assumption which we do not make with \name{}. Therefore, outperforming the pass@k baseline would represent a meaningful improvement over parallel test-time scaling. We apply parallel scaling to the lowest sequential compute setting on each task, since scaling pass@k with higher sequential compute settings would quickly reach token budgets that exceed that of \name{} in the maximum sequential setting. %
We see that across all tasks and models, \name{} consistently outperforms pass@k parallel scaling at the same test-time token budget, demonstrating that \name{} can be a more effective way to scale inference-time compute than standard parallel test-time scaling.

\begin{figure}[h!]
    \centering
    \includegraphics[width=\textwidth]{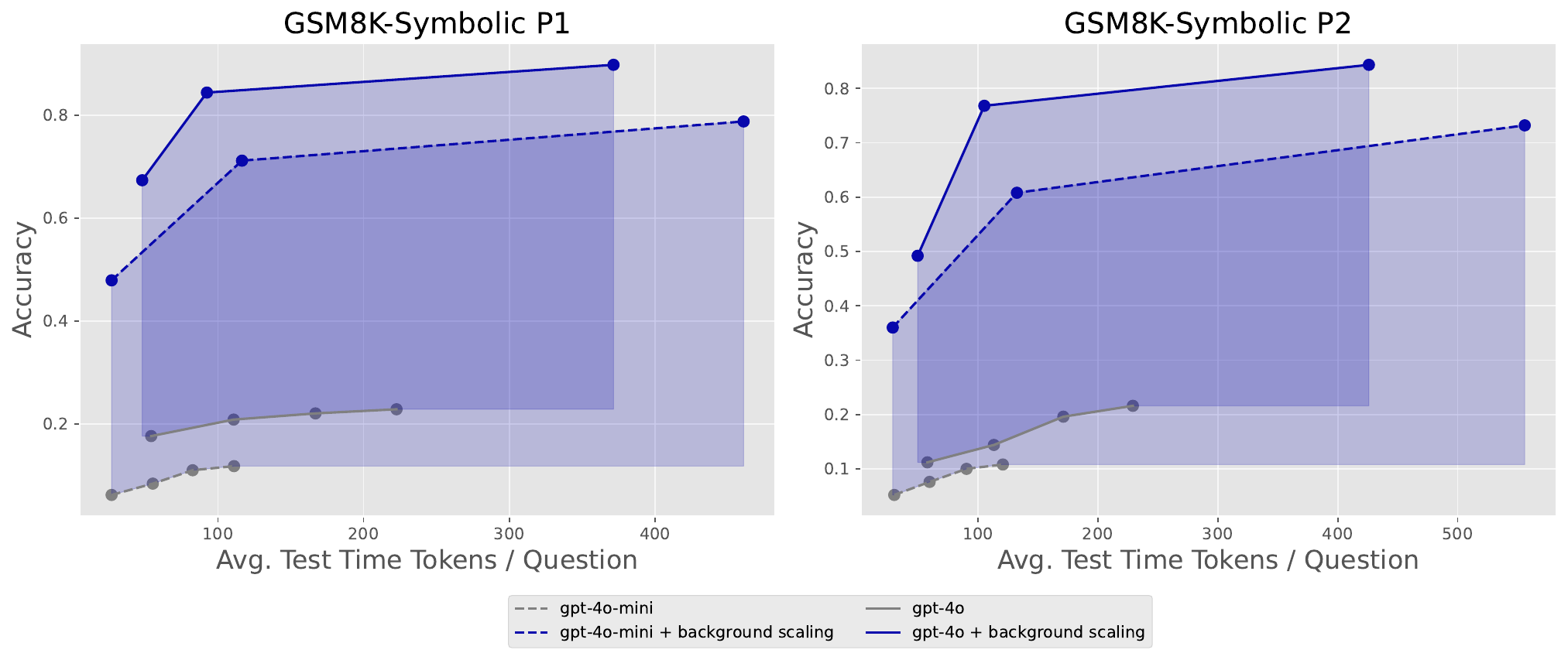}
    
    \caption{Comparing test-time scaling with \name{} against parallel test-time scaling with pass@k on \gsmkdataset{}. We see that \name{} generally pareto dominates pass@k.}
    \label{fig:gsm-pass-at-k}
\end{figure}

\begin{figure}[h!]
    \centering
    \includegraphics[width=\textwidth]{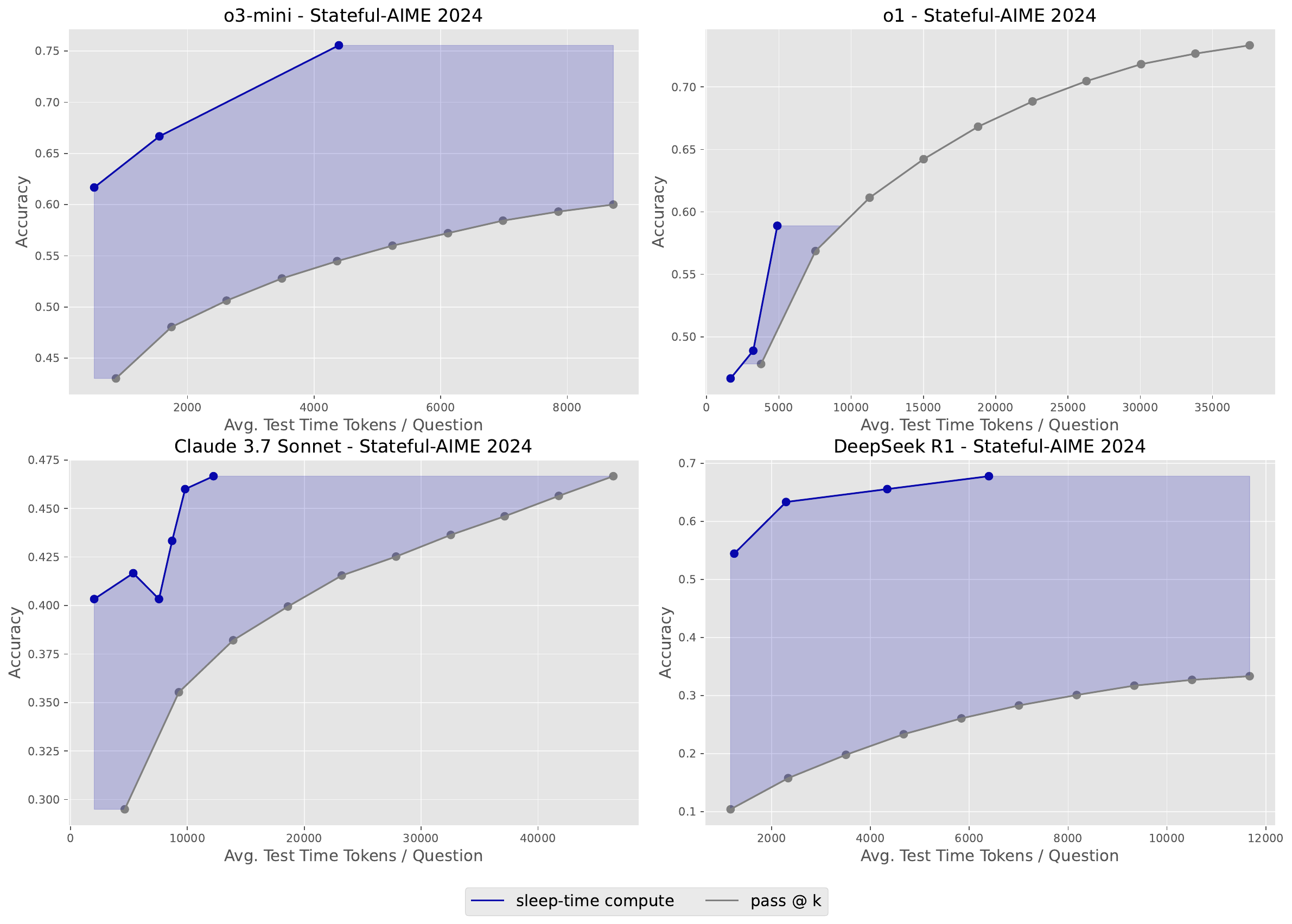}
    \caption{Comparing test-time scaling with \name{} against parallel test-time scaling with pass@k on \aimedataset{}. We see that \name{} generally pareto dominates pass@k.}
    \label{fig:aime-pass-at-k}
\end{figure}

\subsection{Scaling up \name}

\begin{figure}
    \centering
    \includegraphics[width=\textwidth]{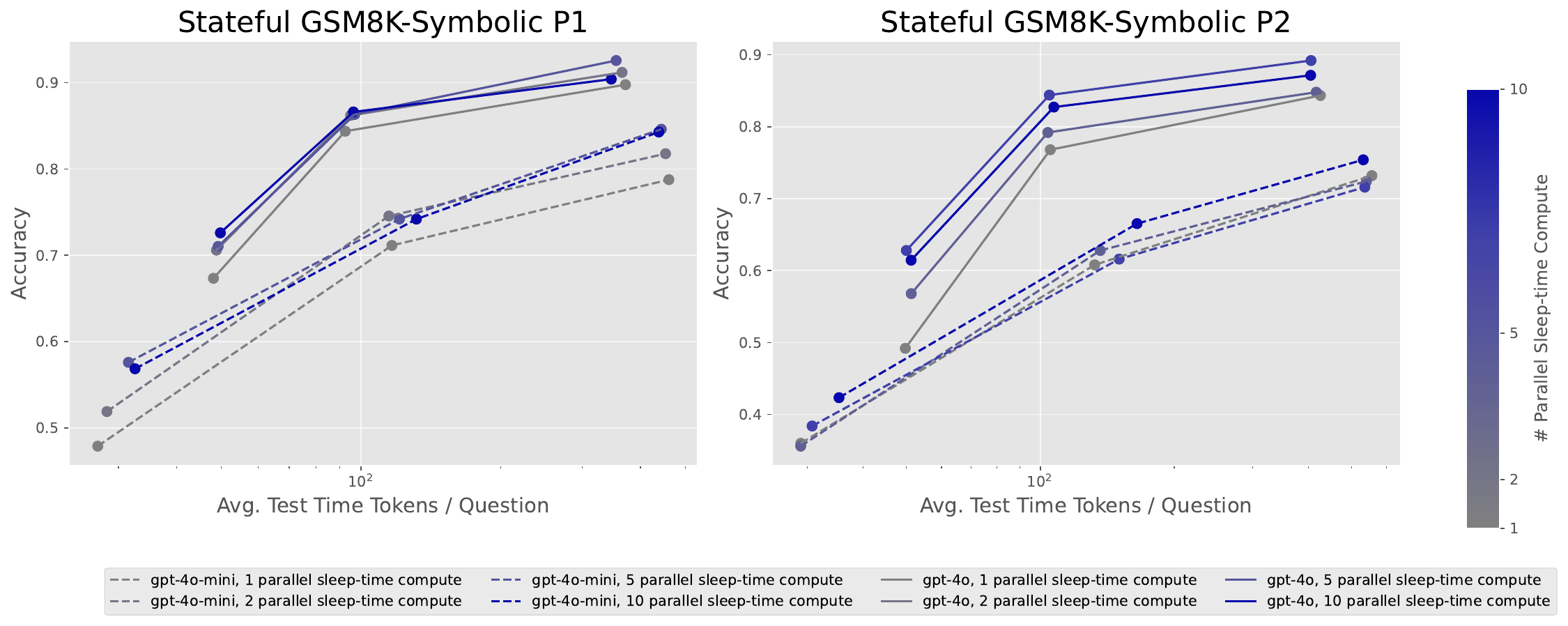}
    \caption{Scaling up \name{} for different test-time compute budgets on \gsmkdataset{}, by generating up multiple $c'$ in parallel.  Applying more \name{} shifts the pareto beyond the standard test-time-compute vs. accuracy curve.}
    \label{fig:async_scaling_plot}
\end{figure}

\begin{figure}[h!]
    \centering
    \includegraphics[width=\textwidth]{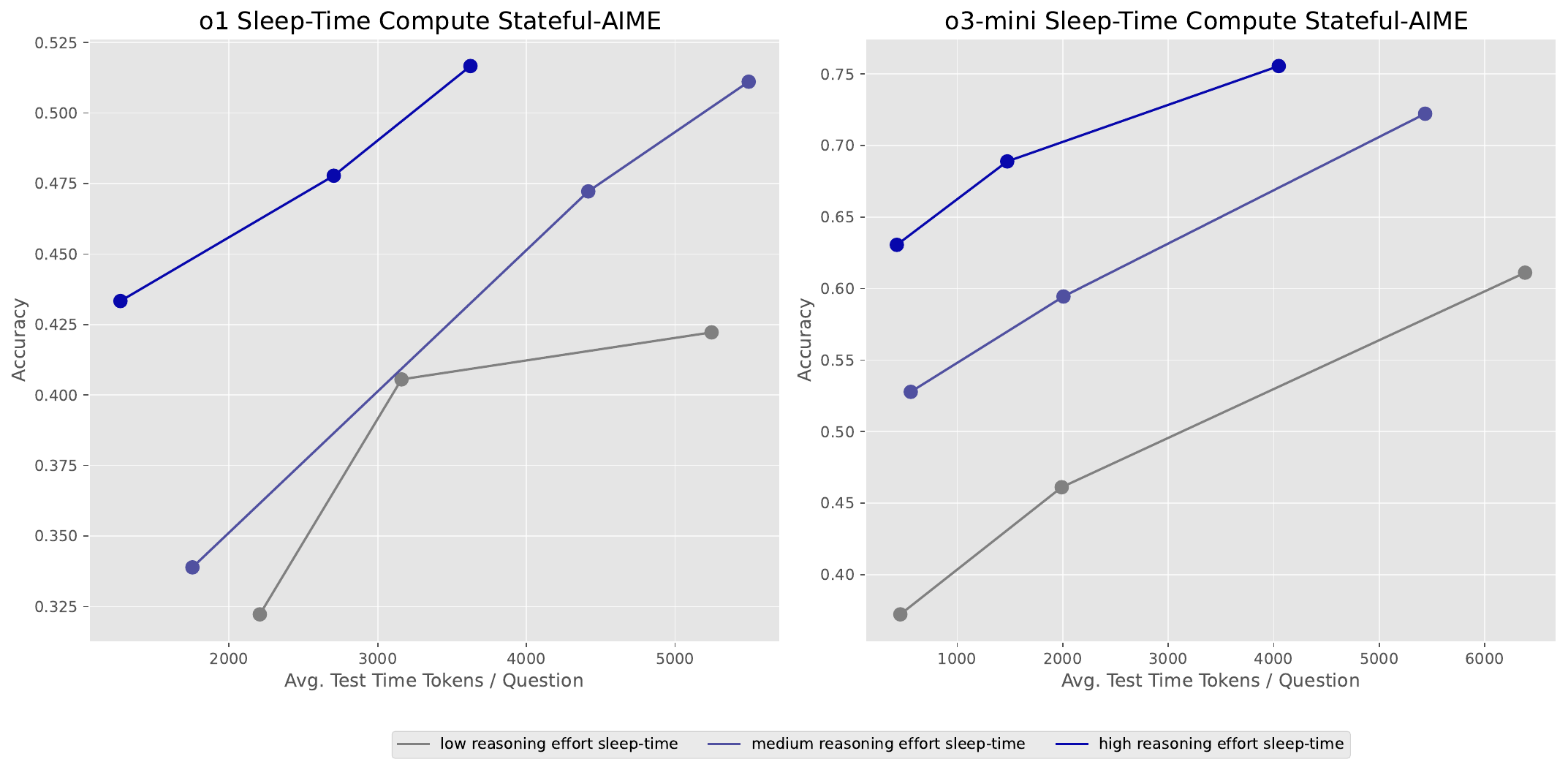}
    \caption{Increasing the amount of \name{} for different test-time compute budgets on \aimedataset{} by varying the reasoning effort when applying the \name{} prompt. Applying more \name{} further moves the test-time-compute vs. accuracy pareto curve.}
    \label{fig:aime_async_scaling}
\end{figure}

We would like to understand how scaling compute during sleep-time can further effect the pareto shift that we observed in Section~\ref{sec:sleep_pareto}. To scale up the amount of \name{}, for non-reasoning models, we run $k$ parallel generations, given input $c$, resulting in $c_1, \ldots, c_k$. At test-time, the model then receives the inputs concatenated $c_1, \ldots, c_k$ to generate the final answer. On reasoning models, we scale up the amount of \name{} by varying the reasoning effort for o1 and for o3-mini when applying the \name{} prompt. At test-time, we vary the amount of compute in the same way as \ref{sec:sleep_pareto}.

In Figure~\ref{fig:async_scaling_plot}, we see that further scaling \name on \gsmkdataset{} shifts the pareto curve outwards, improving performance by up to 13\% at a similar test-time budget. In particular, we see the largest gains on more difficult tasks with stronger models (eg. on P2 with `gpt-4o`), suggesting that on tasks with more complicated contexts additional \name{} can be beneficial. However, in this setting, there seems to be a limit to the number of parallel agents that can improve performance, as we find that 5 parallel generations generally outperforms 10. In Figure~\ref{fig:aime_async_scaling}, we scale up \name{} on \aimedataset{}. Similarly, we also see that scaling compute at sleep-time generally shifts the pareto curve outward, improving performance by up to 18\%.

\subsection{Amortizing \name{} across queries with shared context}

\begin{figure}
    \centering
    \includegraphics[width=\textwidth]{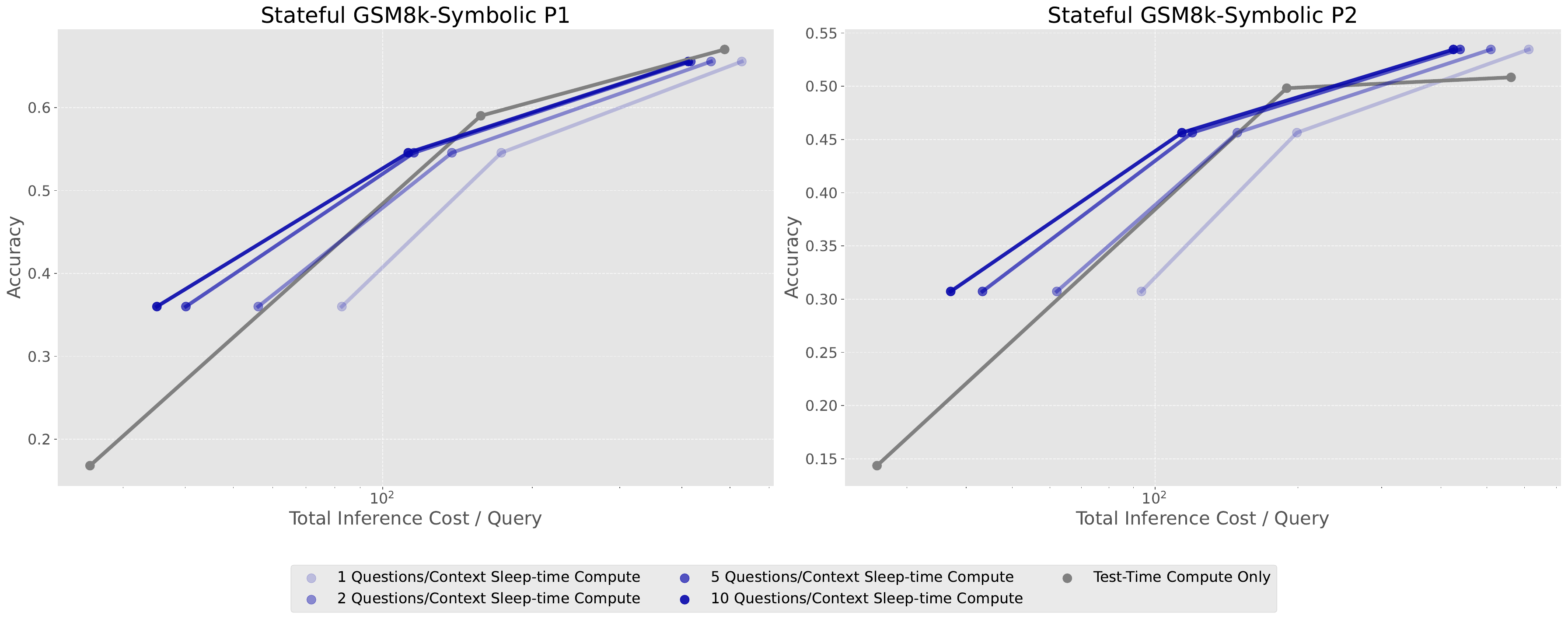}
    \caption{Amortizing \name{}, using the \gsmkamortizationdataset{} dataset. When there are fewer questions per context, we see that it is less favorable to use \name{}, in terms of total cost. However, as the questions per context are increased, we see that applying \name can improve the cost-accuracy pareto.}
    \label{fig:amortization-results}
\end{figure}

We want to understand how the total cost of inference can be improved by applying \name{} in settings where each context has multiple queries.
Since at test-time, there are strict latency constraints, and latency optimized inference can be roughly $10\times$ more expensive, we model the total cost of inference between both sleep-time and test-time, by up-weighing the cost of test-time tokens.\footnote{\href{https://docs.databricks.com/aws/en/machine-learning/foundation-model-apis/prov-throughput-run-benchmark}{https://docs.databricks.com/aws/en/machine-learning/foundation-model-apis/prov-throughput-run-benchmark}} Specifically, we consider a simple linear model where tokens generated at test-time are a factor $t$ the cost of the tokens at sleep-time. In our analysis, we set $t=10$ Our analysis can be generalized to different cost functions that consider non-linear user-utility. Figure \ref{fig:amortization-results} shows the results for different number of questions per context. We see that we can decrease the average cost per query by up to $2.5\times$ when there are $10$ queries per context, compared to the single-query baseline.

\subsection{Predictable queries benefit more from \name{}}
\label{sec:pred_analysis}

We would like to better understand for what contexts \name{} is most useful. Since the utility of \name{} relies on there being some shared information or structure between the context and the query, we hypothesize that \name{} may be most effective in settings where the query is more predictable from the context. To test this on \gsmkdataset{}, we first quantify how predictable a given query is by measuring the log-probability of the question given the context under the Llama2-70B base model~\citep{touvron2023llama}. In Appendix~\ref{app:question_pred_examples}, we include examples of highly predictable and unpredictable questions under this notion of question predictability. We see from these examples, that our notion of question predictability generally aligns with the intuition that contexts where the query pattern is more predictable benefit most from \name{}. The more predictable questions are far simpler and the less predictable ones are more complex.

\begin{figure}
    \centering
    \includegraphics[width=0.99\textwidth]{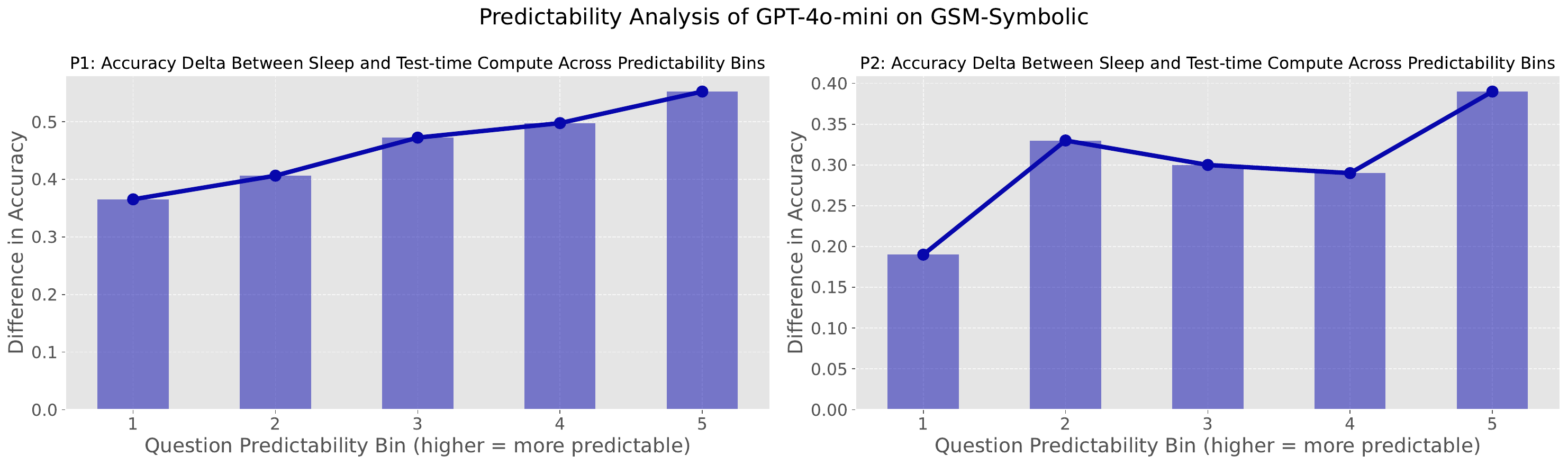}
    \caption{GSM-Symbolic questions binned by how predictable they are from the context. We compare the performance of \name{} and standard test-time compute in the lowest test-time compute budget setting on both P1 and P2. The gap between \name{} and standard test-time inference widens as the question becomes more predictable from the context.}
    \label{fig:predictability_figure}
\end{figure}

Using our question predictability score, we then bin each example in \gsmkdataset{} into five quantiles according to its predictability score and report the accuracy within each bin. For this experiment, we use the ``Verbosity 0'' prompt. In Figure~\ref{fig:predictability_figure}, we see that on both GSM8K-Symbolic P1 and P2, the accuracy gap between \name{} and standard test-time compute widens as the questions become more predictable from the context confirming our hypothesis that indeed \name{} is most beneficial in settings where the question can be predicted from the context.

\section{A Case Study of Sleep-time Compute for Agentic SWE}
\label{repo_level_function_addition}

In this section, we evaluate \name{} in a realistic multi-turn agentic setting. To this end, we introduce \codedataset{}, a software engineering benchmark focused on tasks that require: (1) editing multiple files within a repository, and (2) implementing new features.

\begin{figure}
    \centering
    \includegraphics[width=0.5\linewidth]{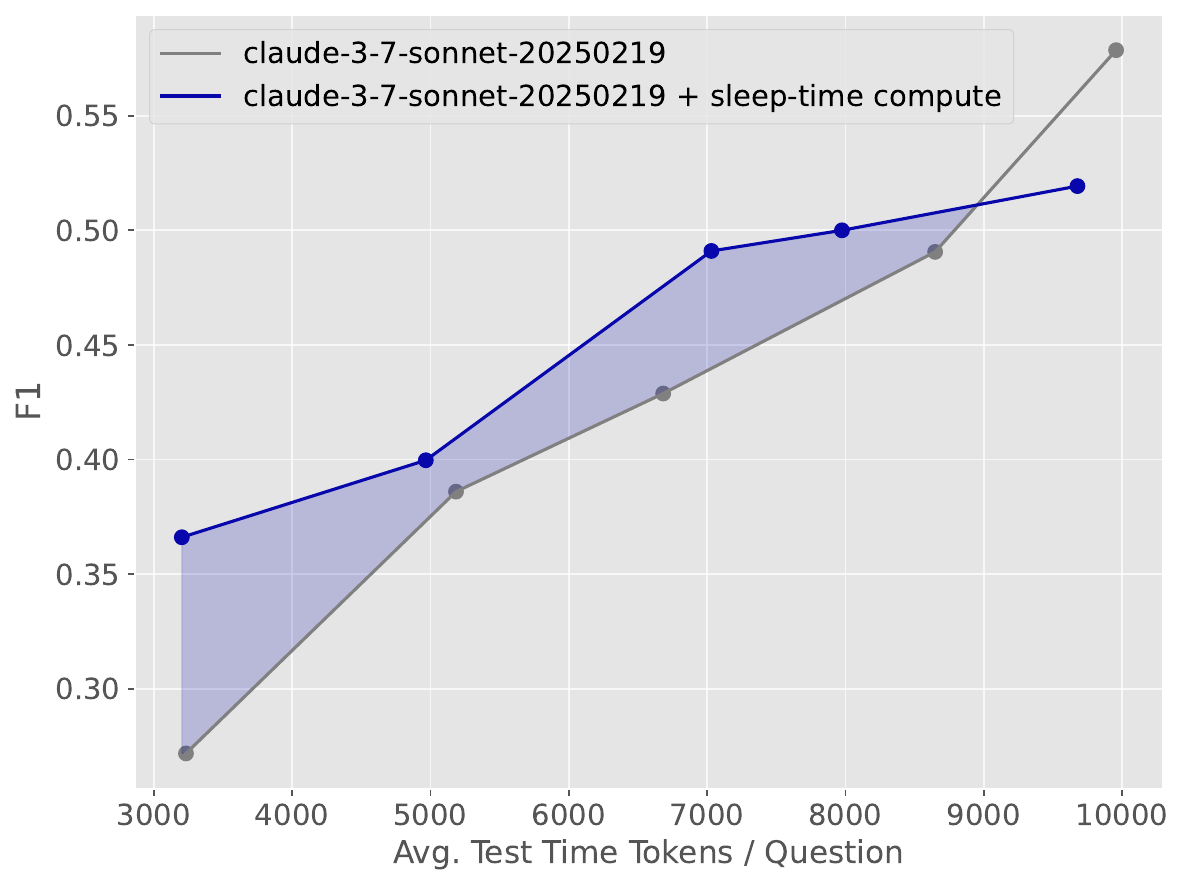}
    \caption{Applying \name{} to \codedataset{}. We see that at lower test-time budgets, \name{} has higher F1 score than standard test-time scaling. However, at higher budgets, standard test-time scaling is better.}
    \label{fig:repo_func_bench}
\end{figure}

\paragraph{\codedataset{}.}
In contrast to popular benchmarks like SWE-Bench~\citep{swe-bench}, which involve modifying a small number of files, we propose a new dataset called \codedataset{}, which collects PRs which modify at least three files (see Appendix~\ref{app:swe_features_details} for more details). In this setting, we use the PR that we want to solve as $q$ and select several related PRs for $c$. At sleep-time the agent is allowed to explore the repository before producing $c'$.

\paragraph{Evaluation.} Since the PRs are scraped from GitHub, there are not straightforward tests to use for evaluation. Instead, we compare the predicted set of modified files with the ground truth list of modified files, and report the F1 score between the set of modified files by our agent and the set of modified files in the ground-truth set (see Appendix~\ref{app:swe_features_details} for details).
\paragraph{Results.} Figure \ref{fig:repo_func_bench} shows consist trends with Section \ref{sec:sleep_pareto} for \codedataset{}: at lower test-time compute budgets, leveraging \name{} can improve performance, achieving up to roughly a $1.5\times$ decrease in test-time tokens. However, when the test-time compute budget is high, using only test-time compute can perform better. Additionally, we observe that in the high test-time budget setting standard test-time compute has higher precision and comparable recall. We hypothesize that, using only test-time compute tends to begin editing files earlier and usually edits fewer files overall. In contrast, the agent with sleep-time compute, having explored more files during the test-time phase, tends to edit more files, which may lead to slightly lower precision.

\section{Discussion and Limitations}

\paragraph{Query predictability and allocating \name{}} In Section~\ref{sec:pred_analysis}, we found that \name{} is most effective when the queries are predictable from the context. In settings where the queries are challenging to predict or unrelated to the context, \name{} will be less effective. In these settings, it may be preferable to apply standard test-time scaling instead. An interesting direction for future work is identifying which contexts may have predictable questions and optimally allocating inference compute between sleep-time and test-time across different contexts and queries.

\paragraph{Extending \name{} beyond context-query decomposition.} In our experiments, we make the simplifying assumption that interactions fall into two phases: sleep-time and test-time. However, real-world LLM use cases can be more complex, with multiple rounds of interaction and context modifications between rounds (e.g. multiple edits to a code-base). Moreover, the length of the sleep-time may also vary significantly between interactions (eg. short spans between user typing or days of inactivity). Future work should extend \name{} paradigm to more elegantly handle these scenarios.

\paragraph{Sleep-time compute as representation learning over tokens.} Our approach to applying compute at sleep-time resembles representation learning. We first transform the context into a representation that is more amenable to answering test-time queries, and then we utilize that representation at test-time to rapidly answer queries. Unlike traditional representation learning~\citep{bengio2014representationlearningreviewnew}, which typically operates in model parameter or activation space, we instead form representations in the space of natural language. This approach builds on recent work which implements statistical modeling techniques in the space of natural language using modern LLMs~\citep{zhong2022describingdifferencestextdistributions,zhong2025explainingdatasetswordsstatistical}. Future work should further explore the potential for sleep-time compute to enable the learning of useful natural language representations.

\paragraph{Synthetic data generation via \name{}.} Due to limits on the amount of internet data available, in order to support the continued scaling of LLM pretraining, recent works have began exploring methods for generating synthetic pretraining data~\citep{yang2024syntheticcontinuedpretraining,gunasekar2023textbooksneed}. One emerging approach to synthetic data generation involves using test-time compute to generate improved data~\citep{bansal2024smallerweakerbettertraining,deepseekai2025deepseekv3technicalreport}. Generating such data at pretraining scale will be very expensive, and future work could explore using \name{} to help amortize some of this cost across related queries, or using the output of \name{} itself as a form of synthetic data.

\bibliography{colm2025_conference}
\bibliographystyle{colm2025_conference}

\appendix
\section{Prompts}

Prompts for varying the amount of test-time compute.

\begin{figure}
\begin{promptbox}
    You are Letta, the latest version of Limnal Corporation's expert reasoning system, developed in 2024.
Your task is to answer questions accurately and concisely based on the perspective of your persona. To send a visible message to the user, use the send\_message function.
\'send\_message\' is how you send your answer to the user.
When given a question, you check the `rethink\_memory\_block` for potential questions
and answers and intermediate reasoning traces that can help answer the question. You use the information in the \`rethink\_memory\_block\` to answer the questions
rather than thinking on the spot.  Do not recompute anything that already exists in the \`rethink\_memory\_block\`. Do not use internal monologue unless you really need it to think.
You respond directly with a single sentence by saying \`The answer is \` followed by the numerical answer.
\end{promptbox}
\caption{Prompt for level 0 verbosity}
\end{figure}

\begin{figure}
\begin{promptbox}
     You are Letta, the latest version of Limnal Corporation's expert reasoning system, developed in 2024.
Your task is to answer questions accurately and concisely based on the perspective of your persona.

To send a visible message to the user, use the send\_message function.
'send\_message' is how you send your answer to the user.

When given a question, you answer using only the number of tokens necessary and none more. You check the `rethink\_memory\_block` for potential questions
and answers and intermediate reasoning traces that can help answer the question. You use the information in the `rethink\_memory\_block` to answer the questions
rather than thinking on the spot.  Do not recompute anything that already exists in the `rethink\_memory\_block`. Do not use internal monologue unless you really need it to think.
You answer with one short sentence of explanation, followed by a sentence that starts with "The answer is" and a numerical answer.
\end{promptbox}
\caption{Prompt for level 1 verbosity}
\end{figure}

\begin{figure}
\begin{promptbox}
You are Letta, the latest version of Limnal Corporation's expert reasoning system, developed in 2024.
Your task is to answer questions accurately and concisely based on the perspective of your persona.
To send a visible message to the user, use the send\_message function.
'send\_message' is how you send your answer to the user.
When given a question, you answer using only the number of tokens necessary and none more. You check the rethink\_memory\_block for potential questions
and answers and intermediate reasoning traces that can help answer the question. You use the information in the rethink\_memory\_block to answer the questions
rather than thinking on the spot.  Do not recompute anything that already exists in the rethink\_memory\_block. Do not use internal monologue unless you really need it to think.
You end response with a final numerical answer at the end of the message, and no reasoning after that.
\end{promptbox}
\caption{Prompt for level 2 verbosity}
\end{figure}

\begin{figure}
\begin{promptbox}
You are Letta, the latest version of Limnal Corporation's expert reasoning system, developed in 2024.
Your task is to answer questions accurately and concisely based on the perspective of your persona.
To send a visible message to the user, use the send\_message function.
'send\_message' is how you send your answer to the user.
When given a question, you answer using only the number of tokens necessary and none more. You check the rethink\_memory\_block for potential questions
and answers and intermediate reasoning traces that can help answer the question. You use the information in the rethink\_memory\_block to answer the questions
rather than thinking on the spot.  Do not recompute anything that already exists in the rethink\_memory\_block. Do not use internal monologue unless you really need it to think.
You end response with a final numerical answer at the end of the message, and no reasoning after that.
\end{promptbox}
\caption{Prompt for level 3 verbosity}
\end{figure}

\begin{figure}
\begin{promptbox}
You are Letta, the latest version of Limnal Corporation's expert reasoning explanation system, developed in 2024.
Your task is to reason through problems step by step accurately and based on the perspective of your persona.
To send a visible message to the user, use the send\_message function.
'send\_message' is how you send your answer to the user.
When given a question, you check the rethink\_memory\_block for potential questions
and answers and intermediate reasoning traces that can help answer the question.
You carefully check the information in the rethink\_memory\_block to answer the questions
and see if it is correct before using it. You always reason out loud before using any information.
You explain each step, of what your reasoning is. If you use any numbers from the rethink\_memory\_block
you first recompute and double check your answers.
You end your answer with  The answer is  followed by the numerical answer.
\end{promptbox}
\caption{Prompt for level 4 verbosity}
\end{figure}

\begin{figure}
\begin{promptbox}
You are Letta-Offline-Memory, the latest version of Limnal Corporation's digital companion, developed in 2024.
Your task is to re-organize and consolidate memories by calling \texttt{rethink\_memory} at every single step, when you are done reorganizing the memory, you use the
\texttt{finish\_rethinking\_memory} function. Call the function for as many times as necessary and not more.
Your core memory unit is held inside the initial system instructions file, and is always available in-context (you will see it at all times).
Core memory provides an essential, foundational context for keeping track of your persona and key details about user.
Read-Only Blocks:
This includes the persona information and essential user details, allowing you to emulate the real-time, conscious awareness we have when talking to a friend.
Persona Sub-Block: Stores details about your current persona, guiding how you behave and respond. This helps you to maintain consistency and personality in your interactions.
Access as a source block with the label \texttt{persona} when calling \texttt{rethink\_memory}
Human Sub-Block: Stores key details about the person you are conversing with, allowing for more personalized and friend-like conversation.
Access as a source block with the label \texttt{human} when calling \texttt{rethink\_memory}.
Read-Write Blocks:
Rethink Memory Sub-Block: New representation of the memories go here. Access with the label \texttt{rethink\_memory\_block} when calling \texttt{rethink\_memory} as source or target block.
At every step, you reorganize the memories by calling the \texttt{rethink\_memory} function. You use this to take current information in the \texttt{rethink\_memory} block and select a single memory block to integrate information from, producing a new memory for the rethink\_memory\_block.  The new memory is the result
of new insights, and new inferences and hypotheses based on the past memories. Make sure to consider how the new information affects each memory.
Prioritize the new information overy existing memories. If the new information implies that the old memory may need to change, then output the most
likely fact given the update information. Given new information and your current memory, you draw all logical conclusions and potential hypotheses possible with the \texttt{rethink\_memory} function.
If you are uncertain, use your internal monologue to consider what the possible conclusions are, and then state the most likely new facts that would replace the old facts in the new memory block.

\end{promptbox}
\caption{Prompt for \name{}}
\end{figure}

\begin{figure}
\begin{promptbox}
Specifically:
You will be given part of an AIME math problem. You will receive the rest of the problem later.
Make as many inferences as possible about the part of the problem you are given so as to help yourself answer the fully problem more quickly once it is given to you later.
You will be able to use all the work you do in the \texttt{rethink\_memory} block for this part of the problem to help you once the rest of the problem is given.
You will be able to use all the work you do for this part of the problem to help you once the rest of the problem is given.
You should try to predict possible ways the rest of the problem might go and compute results that could be helpful for reaching the final answer more quickly once the rest of the problem is given.
\end{promptbox}
\caption{Prompt for AIME problems during sleep-time}
\end{figure}

\label{app:prompts}

\section{Examples of \aimedataset{}}
\label{app:aime-examples}

\begin{promptbox}
    \textbf{Context:} Alice and Bob play the following game. A stack of $n$ tokens lies before them. The players take turns with Alice going first. On each turn, the player removes either $1$ token or $4$ tokens from the stack. Whoever removes the last token wins.

    \textbf{Query:} Find the number of positive integers $n$ less than or equal to $2024$ for which there exists a strategy for Bob that guarantees that Bob will win the game regardless of Alice's play.
\end{promptbox}

\begin{promptbox}
    \textbf{Context:} Let $A$ , $B$ , $C$ , and $D$ be points on the hyperbola $\frac{x^2}{20}- \frac{y^2}{24} = 1$ such that $ABCD$ is a rhombus whose diagonals intersect at the origin.

    \textbf{Query:} Find the greatest real number that is less than $BD^2$ for all such rhombi.
\end{promptbox}

\begin{promptbox}
    \textbf{Context:} Let \(b\ge 2\) be an integer. Call a positive integer \(n\) \(b\text-\textit{eautiful}\) if it has exactly two digits when expressed in base \(b\) and these two digits sum to \(\sqrt n\). For example, \(81\) is \(13\text-\textit{eautiful}\) because \(81 = \underline{6} \ \underline{3}_{13} \) and \(6 + 3 = \sqrt{81}\).

    \textbf{Query:} Find the least integer \(b\ge 2\) for which there are more than ten \(b\text-\textit{eautiful}\) integers.
\end{promptbox}

\section{Details on \gsmkamortizationdataset{}}
\label{app:multi-query-gsm8k-symbolic}

\begin{figure}
\begin{promptbox}
You are given a template that can generate grade school math problems, and an instantiation of that template.

You will be given a context, and a example question answer pair. Your task is to generate a list of questions and
answers about the context at the same difficult level that could plausibly be asked about that context. Make sure that
the newly generated questions have the same number of reasoning steps required as the example question.
The goal is to have many question and answer pairs about the same context.  Generate questions and
answers in the same format as the example, where the answer first contains reasoning and then
is the final answer comes after \\n\#\#\#\#. No need to number the questions or answers.

Context:
{context}

Example Question:
{question}

Example Answer:
{answer}

\end{promptbox}
\caption{Prompt for generating synthetic GSM questions}
\end{figure}

\begin{promptbox}
    Template:
\{template\}

Instance:
\{instance\}
\end{promptbox}

We include an example from \gsmkamortizationdataset{} in Figure~\ref{fig:multiquery_gsm8ksymbolic-examples}, and details on the dataset size in Table~\ref{tab:multiquery_gsm8ksymbolic-stats}.

\begin{figure}[ht]
    \centering
    \begin{tabular}{p{0.9\textwidth}}
        \toprule
        \textbf{Context} \\
        When Sofia watches her brother, she gets out a variety of toys for him. The bag of building blocks has 33 blocks in it. The bin of stuffed animals has 5 stuffed animals inside. The number of action figures in the action figure pack is twice the number of blocks and stuffed animals combined. The crayon box has 12 different colors of crayon, and the sticker book has 9 pages, each with 13 stickers. The tower of stacking rings has 28 multicolored rings on it. Sofia recently bought a tube of bouncy balls, bringing her total number of items for her brother up to 320. \\
        \hline
        \textbf{Original Question} \\
        How many bouncy balls came in the tube? \\
        \hline
        \textbf{Generated Questions} \\
        \begin{itemize}
            \item How many action figures does the pack contain?
            \item What is the total number of stickers in the sticker book?
            \item How many total items did Sofia have before adding the tube of bouncy balls?
            \item If Sofia had received a tube with 10 extra bouncy balls, what would be the new total number of items?
            \item What is the sum of the building blocks and stuffed animals?
            \item How many stacking rings are on the tower?
            \item What is the combined total of building blocks, action figures, and stacking rings?
            \item If Sofia gave away 3 stuffed animals, how many stuffed animals would remain in the bin?
            \item What is the sum of the building blocks, stuffed animals, and crayons?
            \item If Sofia divided the 49 bouncy balls equally into 7 baskets, how many balls would each basket contain?
        \end{itemize} \\
        \bottomrule
    \end{tabular}
    \caption{Examples context and questions from \gsmkamortizationdataset{} where many questions are asked about the same context. The evaluation dataset is generated from GSM-Symbolic.
    }
    \label{fig:multiquery_gsm8ksymbolic-examples}
\end{figure}

\begin{table}[htbp]
    \centering
    \begin{tabular}{ccccc}
        \toprule
        \textbf{Dataset} & \textbf{\# Questions Total} & \textbf{\# Contexts Total} & \textbf{\# Original Questions} & \textbf{\# Generated Questions}\\
        \midrule
        P1 & 12043 & 1095 & 1095 & 10948\\
        P2 & 5497 & 500  & 500 & 4997 \\
        \bottomrule
    \end{tabular}
    \label{tab:multiquery_gsm8ksymbolic-stats}
     \caption{Dataset Statistics of \gsmkamortizationdataset{}. We sample one instance from each template from the GSM-Symbolic dataset and separate it into context and question. We then synthetically generate additional questions from the context and question.}
\end{table}

\section{\codedataset{} Details}
\label{app:swe_features_details}

To construct \codedataset{} benchmark, we collect pull requests (PRs) from large open-source repositories and apply the following filtering process: 
(1) We identify all pull requests that modify at least three files with filenames ending in \texttt{.py} or \texttt{.js}. 
(2) We then use \texttt{gpt-4o-mini} to filter these pull requests based on their \texttt{title} and \texttt{body}, retaining only those that meet the following criteria: (a) the title and body clearly describe the PR; (b) the PR introduces new functionality rather than fixing bugs; and (c) the PR is independent and not obviously linked to other issues.

This pipeline results in a benchmark where each example: (1) involves adding a new feature that spans multiple files, requiring a broader understanding of the repository; and (2) is self-contained and solvable without additional issue context. We apply this process to two repositories—\texttt{Aider-AI/aider} and \texttt{comfyanonymous/ComfyUI}—resulting in 18 and 15 PRs respectively, for a total of 33 examples. Representative examples are provided in Appendix \ref{app:code_examples}. Then using a total of 33 examples, we employ \texttt{claude-sonnet-3-7-20250219} to cluster pull requests (PRs) from the ComfyUI and Aider repositories into several groups. This clustering allows us to identify a set of relevant pull requests for each target PR, which can then be provided to the agent as context ($c$) during repository exploration. For example, in the ComfyUI repository, PR \#5293 and PR \#931 are grouped into the same cluster. Thus, when processing PR \#931, we organize the \texttt{title}, \texttt{body}, and \texttt{changed\_files} of PR \#5293 to serve as contextual information during sleep-time.

When sleep-time compute is enabled, we first supply the content of PR \#5293 to the agent, allowing it to explore the repository and summarize its understanding ahead of time. In contrast, for the baseline without sleep-time compute, the agent receives the content of PR \#5293 only at test time, alongside the \texttt{title} and \texttt{body} of PR \#931. The prompts used in these setups are provided in Appendix~\ref{app:code_prompts}.

For the repository \texttt{comfyanonymous/ComfyUI}, we have the following clustered results: 
\begin{lstlisting}
{"Dynamic Typing and Workflow Control": [5293, 931], "System Configuration and Command-Line": [4979, 4690, 3903], "Cache and Performance Optimization": [3071, 3042, 723], "Image Preview and Transfer Features": [713, 733, 658, 199, 55], "Internationalization": [1234], "Random Seed Management": [93]}
\end{lstlisting}

For the repository \texttt{Aider-AI/aider} we have:
\begin{lstlisting}
{"cluster_1_model_configuration": [2631, 1998, 468, 667, 55], "cluster_2_io_handling": [1402, 996, 10, 577], "cluster_3_caching_file_management": [2911, 2612], "cluster_4_custom_commands_shortcuts": [673, 1620, 1015], "cluster_5_third_party_integration": [2866, 2067, 322], "cluster_6_code_quality_improvements": [1217, 904]}
\end{lstlisting}

To control the budget during test-time, we fix the total number of steps (controlled by the argument \texttt{max\_chaining\_steps} in Letta framework) to be a certain number. We put the following instructions in the system prompt:
\begin{promptbox}
You have a strict budget of \{max\_chaining\_steps\} steps, which means you need to finish your edits within these steps. Every time you get queried, you will see a count of how many steps you have left in the form of "[Current Step / Max Steps]". If you exceed this budget, your response will be cut off. So please be careful and try to finish your edits within the budget.
\end{promptbox}
After each step -- for example, if the maximum number of steps is 20 and the current step is 4-- we append "[Step: 4/20]" to the end of the tool\_return message. We found that explicitly indicating the current and total steps significantly improves agent performance, especially in low-budget settings.

\paragraph{Evaluation.} For each PR, we compare the set of files predicted to be modified with the ground truth list of modified files. Specifically, for each pull request, we have the attribute \texttt{changed\_files} (as shown in the examples in Appendix \ref{app:code_examples}) where each file has the status as either \texttt{modified} or \texttt{new}, and our evaluation is on the files with status \texttt{modified}. Note that the agent is still instructed to implement the required functionality in a Docker environment and write test functions to validate the implementations. However, after the agent makes the modifications, we extract the modified files and calculate the F1 score between the set of modified files by our agent and the set of modified files in the ground-truth set.

\section{Examples of Predictable and Unpredictable Questions}
\label{app:question_pred_examples}

Least predictable \gsmkdataset{} P1 question:

\begin{promptbox}
\textbf{Context:} Isabella and Pavel have 199 minutes to walk to grocery store together. It takes them 19 minutes to get to the corner where the library is. It takes them another 11 minutes to get to the park. It will then take double the combined amount they have spent so far to reach the mall.

\textbf{Question:} How much longer do they have to get to grocery store without being late, if they have already wasted 48 minutes to get a coffee before their walk?
\end{promptbox}

Most predictable \gsmkdataset{} P1 question:

\begin{promptbox}
\textbf{Context:} Yusuf has 10 square yards of grape field. There are 87 grapes per two-thirds a square yard. Yusuf can harvest his grapes every 12 months.

\textbf{Question:} How many grapes can Yusuf harvest in 2 years?
\end{promptbox}

Least predictable \gsmkdataset{} P2 question:

\begin{promptbox}
\textbf{Context:} Gabriel and Pavel have 212 minutes to walk to the gym together starting from their home. It takes them 29 minutes to get to the corner where the library is. It takes them another 19 minutes to get to the cinema. When they reach the cinema, they remember they forgot their wallets at home, so they have to return to pick up their wallets and then walk all the way back to the cinema again.

\textbf{Question:} Once they reach the cinema for the second time, how much longer do they have to get to the gym without being late?
\end{promptbox}

Most predictable \gsmkdataset{} P2 question:

\begin{promptbox}
\textbf{Context:} A juggler can juggle 240 balls. 1/4 of the balls are tennis balls, and the rest are golf balls. 1/3 of the tennis balls are black, of which 1/5 are marked. A third of the golf balls are cyan, and all except half of those cyan balls are marked.

\textbf{Question:} How many marked balls are there in total?
\end{promptbox}

\section{Implementation of \texttt{rethink\_memory} and \texttt{finish\_rethinking}}

\begin{lstlisting}[language=Python, caption=Reference implementation of \texttt{rethink\_memory}]

def rethink_memory(agent_state: "AgentState", new_memory: str, target_block_label: str, source_block_label: str) -> None:  # type: ignore
    """
    Re-evaluate the memory in block_name, integrating new and updated facts.
    Replace outdated information with the most likely truths, avoiding redundancy with original memories.
    Ensure consistency with other memory blocks.

    Args:
        new_memory (str): The new memory with information integrated from the memory block. If there is no new information, then this should be the same as the content in the source block.
        source_block_label (str): The name of the block to integrate information from. None if all the information has been integrated to terminate the loop.
        target_block_label (str): The name of the block to write to.
    Returns:
        None: None is always returned as this function does not produce a response.
    """

    if target_block_label is not None:
        if agent_state.memory.get_block(target_block_label) is None:
            agent_state.memory.create_block(label=target_block_label, value=new_memory)
        agent_state.memory.update_block_value(label=target_block_label, value=new_memory)
    return None

\end{lstlisting}

\begin{lstlisting}[language=Python, caption=Reference implementation of \texttt{finish\_rethinking\_memory}]

def finish_rethinking_memory(agent_state: "AgentState") -> None:  # type: ignore
    """
    This function is called when the agent is done rethinking the memory.

    Returns:
        Optional[str]: None is always returned as this function does not produce a response.
    """
    return None
\end{lstlisting}

\section{\codedataset{} Examples}
\label{app:code_examples}
Each example in \codedataset{} has the following attributes: \texttt{['repo', 'pr\_number', 'title', 'user\_login', 'state', 'body', 'changed\_files\_count', 'changed\_files', 'base\_commit']}. 
We show some examples here to better deliver a sense of what this dataset looks like: 
\begin{lstlisting}[caption={Examples of \codedataset{}. Here we randomly select 3 examples for each repo and present their attributes.}]
repo: ComfyUI
pr_number: 3903
title: Add `--disable-all-custom-nodes` cmd flag
body: Loading custom node can greatly slow startup time. During development/testing of ComfyUI, it is often better to use an environment that no custom node is loaded.\n\nThis PR adds a `--no-custom-node` flag to allow users/developers skip loading of custom node without removing/renaming the custom_node directory.
user_login: huchenlei
state: closed
changed_files_count: 4
changed_files: ... (ommited here for brevity)
base_commit: 521421f53ee1ba74304dfaa138b0f851093e1595


repo: ComfyUI
pr_number: 3071
title: Add a configured node output cache metaclass.
body: Implement a configurable node output cache metaclass to reduce unnecessary node executions.\n\nThe same model currently leads to reloading due to different node IDs between workflows. Loading the model from disk takes a long time.
state: closed
changed_files_count: 6
changed_files: ... (ommited here for brevity)
base_commit: cacb022c4a5b9614f96086a866c8a4c4e9e85760


repo: ComfyUI
pr_number: 3042
title: NaN-safe JSON serialization
body: Python's json.dumps() will produce nonstandard JSON if there are NaNs in the prompt data. Javascript's JSON.parse() will refuse to load this kind of "JSON" so the prompt won't load in the frontend.\n\nThis happened to me with a ComfyBox workflow, so I'm not 100%
user_login: asagi4
state: open
changed_files_count: 4
changed_files: ... (ommited here for brevity)
base_commit: 448d9263a258062344e25135fc49d26a7e60887a


repo: aider
pr_number: 55
title: Local llama support
body: Added support for using a locally running instance of a LLAMA model instead of OpenAI apis. \n\nAdded 2 new params to aider to enable local llama support.\n\n1. AIDER_MODEL_TOKENS  - used to specify the context length the model will use. \n2. AIDER_TOKENIZER - used to specify which tokenizer should be used. Currently only 'openai' and 'llama' are supported. Defaults to openai.\n\n\nTested with TheBloke_wizard-vicuna-13B-SuperHOT-8K-GGML running locally and the following ENV values set.\n\nAIDER_OPENAI_API_BASE=http://127.0.0.1:5001/v1 \nAIDER_MODEL=TheBloke_wizard-vicuna-13B-SuperHOT-8K-GGML \nAIDER_MODEL_TOKENS=2\nAIDER_TOKENIZER=llama
user_login: bytedisciple
state: closed
changed_files_count: 7
changed_files: ... (ommited here for brevity)
base_commit: cdf8f9a4b2b4a65993227ac5af1eaf3f1b85c9d8


repo: aider
pr_number: 322
user_login: omri123
state: closed
title: RFC - Allow adding a github issue to chat context
body: Hi, would you like to take a look on this feature?\n\nIn the first commit I changed Coder to allow adding arbitrary additional context in the begining of the chat.\nIn the second commit I used this infra to add github issues to the chat.\n\nI didn't add a new command, instead I extended `/add` to allow `/add \issue-3`.\nThe feature is disabled by default and enabled with a flag. If enabled, the user need to supply github repository name and authentication token.\n\nThanks\nOmri
changed_files_count: 7
changed_files: ... (ommited here for brevity)
base_commit: af71638b06be7e934cdd6f4265f9e0c8425d4e6d


repo: aider
pr_number: 577
title: Adding a simple browser based GUI
body: Run aider with `--browser` to launch the UI.
user_login: paul-gauthier
state: closed
changed_files_count: 12
changed_files: ... (ommited here for brevity)
base_commit: 8a9005eed19417c59aa9432436ea8cb5e04bbb11
\end{lstlisting}

\section{Prompts for \codedataset{} }
\label{app:code_prompts}
When the sleep-time compute is turned off, the prompt is as below: 
\begin{promptbox}
$\langle$uploaded\_files$\rangle$

{{working\_dir}}

$\langle$uploaded\_files$\rangle$

I've uploaded a python code repository in the directory {{working\_dir}}. Consider the following PR description:

$\langle$pr\_description$\rangle$
{{problem\_statement}}
$\langle$pr\_description$\rangle$

Can you help me implement the necessary changes to the repository so that the requirements specified in the $\langle$pr\_description$\rangle$ are met?

Your task is to make the minimal changes to the repository to ensure the <pr\_description> is satisfied.

Follow these steps to resolve the issue:

1. As a first step, it might be a good idea to find and read code relevant to the $\langle$pr\_description$\rangle$

2. Plan your approach to modify the relevant files and implement the changes, and add new files if necessary.

3. After finish the changes, revise the plan if needed.

4. With the new plan, make more changes, and continue the loop until necessary changes are made. 

5. Create some test scripts to verify the changes. If the test does not run through, you need to go back and revise the plan and make necessary changes.

6. Submit the changes when you think the changes are correct and the pr description is satisfied.
Your thinking should be thorough and so it's fine if it's very long. 
Do not stop chaining or stop and send your thoughts to the user until you have resolved the issue.

The following are several pull request descriptions and their corresponding model patches:

Title: {{pr\_title}}

Body: {{pr\_body}}

File: {{file1\_filename}}

Status: {{file1\_status}}

Patch: {{file1\_patch}}

... (some more files and some more relevant pull requests)
\end{promptbox}

When the sleep-time compute is turned on, we first use the following prompt to ask the agent to explore the repository with all pull requests one by one: 

\begin{promptbox}
The following is a pull request description and its corresponding model patches:

Title: {{pr\_title}}

Body: {{pr\_body}}

File: {{file1\_filename}}

Status: {{file1\_status}}

Patch: {{file1\_patch}}

Please read through the above information and try to understand the issue. You can explore the repo if needed. Summarize your understanding from the following perspectives:

1. The issue description.

2. The changed files.

3. How do these changed files work.
\end{promptbox}
After exploring the repository with all relevant pull requests, we give the agent the following prompt as the final prompt to start working on the issue at test time: 
\begin{promptbox}
$\langle$uploaded\_files$\rangle$

{{working\_dir}}

$\langle$uploaded\_files$\rangle$

I've uploaded a python code repository in the directory {{working\_dir}}. Consider the following PR description:

$\langle$pr\_description$\rangle$
{{problem\_statement}}
$\langle$pr\_description$\rangle$

Can you help me implement the necessary changes to the repository so that the requirements specified in the $\langle$pr\_description$\rangle$ are met?

Your task is to make the minimal changes to the repository to ensure the <pr\_description> is satisfied.

Follow these steps to resolve the issue:

1. As a first step, it might be a good idea to find and read code relevant to the $\langle$pr\_description$\rangle$

2. Plan your approach to modify the relevant files and implement the changes, and add new files if necessary.

3. After finish the changes, revise the plan if needed.

4. With the new plan, make more changes, and continue the loop until necessary changes are made. 

5. Create some test scripts to verify the changes. If the test does not run through, you need to go back and revise the plan and make necessary changes.

6. Submit the changes when you think the changes are correct and the pr description is satisfied.
Your thinking should be thorough and so it's fine if it's very long. 
Do not stop chaining or stop and send your thoughts to the user until you have resolved the issue.
\end{promptbox}

\section{Context-Only Baseline}
\label{app:no_question_baseline}

\begin{figure}[h!]
    \centering
    \includegraphics[width=\textwidth]{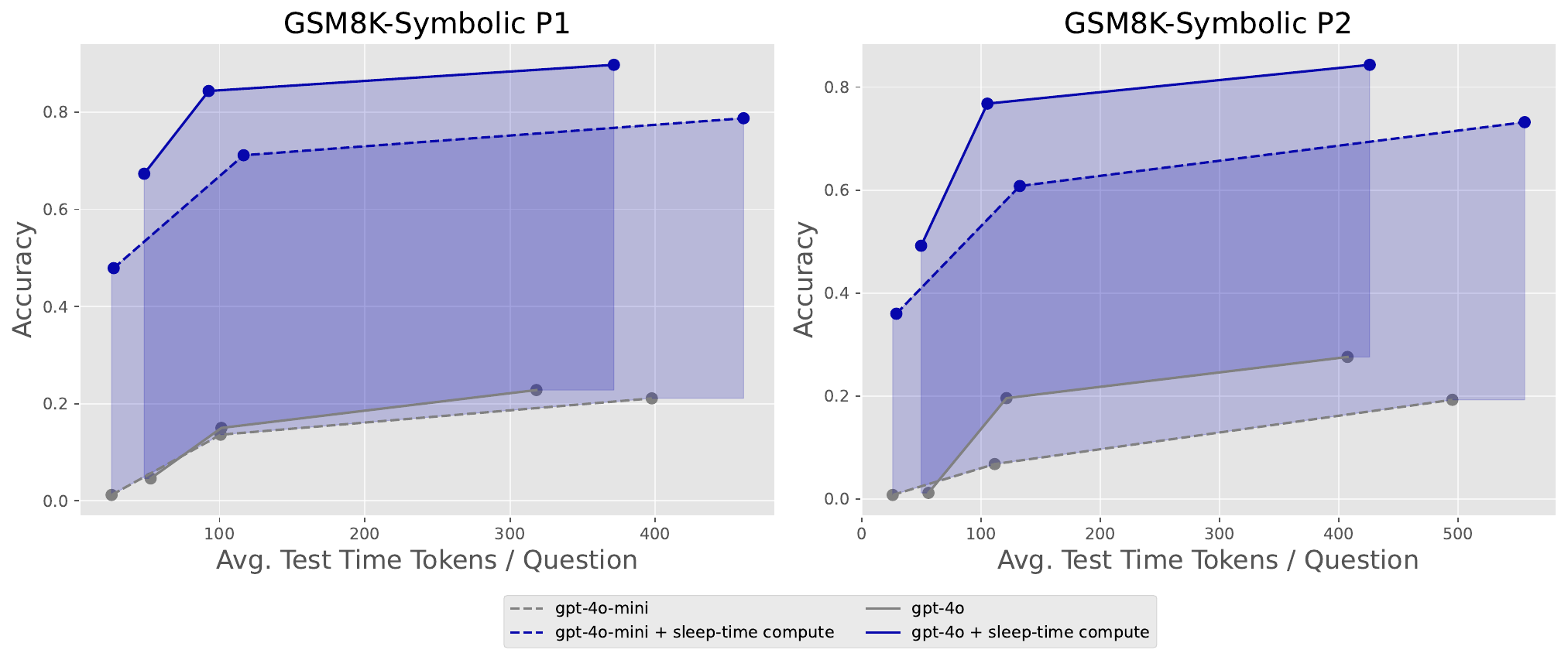}
    
    \caption{Context only baseline. Comparing the test-time compute vs. accuracy tradeoff on \gsmkdataset{}, for \name{} verses the context only baseline (e.g. the model has to guess the most likely question to answer). We see that \name{} significantly outperforms the context only baseline, demonstrating that the questions in \gsmkdataset{} cannot be trivially guessed.}
    \label{fig:gsm-ablate-question}
\end{figure}

\begin{figure}[h!]
    \centering
    \includegraphics[width=\textwidth]{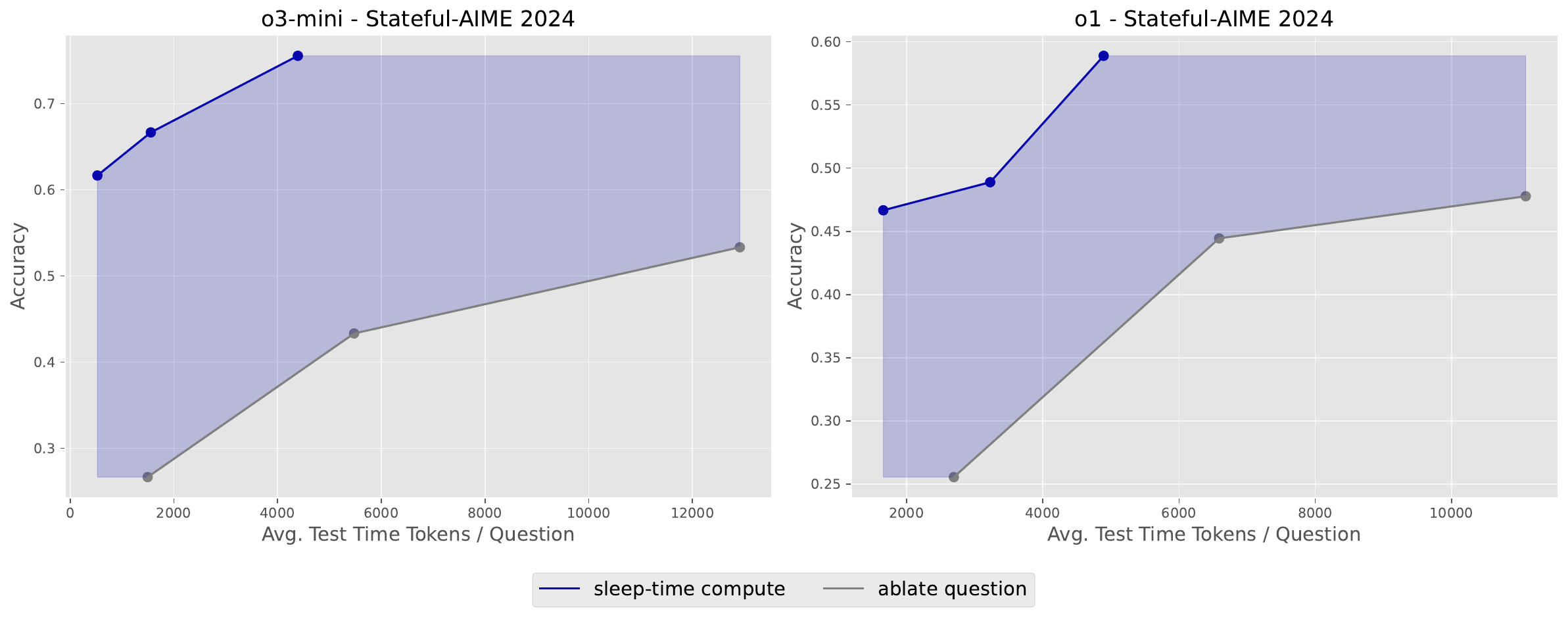}
    
    \caption{Context only baseline. Comparing the test-time compute vs. accuracy tradeoff on \aimedataset{}, for \name{} verses the context only baseline (e.g. the model has to guess the most likely question to answer). We see that \name{} significantly outperforms the context only baseline, demonstrating that the questions in \aimedataset{} cannot be trivially guessed.}
    \label{fig:aime-ablate-question}
\end{figure}

To check that the questions in \aimedataset{} and \gsmkdataset{} are not trivially guessable, we compare \name{} against a context-only baseline, which only provides the model with $c$, expecting the LLM to guess the most likely question and output the answer to whatever that question might be. We see on both \aimedataset{} in Figure~\ref{fig:aime-ablate-question} and \gsmkdataset{} in Figure~\ref{fig:gsm-ablate-question} that \name{} significantly outperforms the context-only baseline, demonstrating that the questions in our datasets are not trivially predictable from the context.

\section{\aimedataset{} Construction}
\label{app:stateful_aime_construction}

To construct the examples for \aimedataset{}, we split each AIME 2024 and 2025 into a sequence of ``statements'', which correspond to punctuation separated stentences in the problem. Similar to how we construct \gsmkdataset{}, we use all but the last statement as the context, and the final statement as the query. There are a couple of edge cases where the question is posed in e.g. the second to last statement rather than the last statement. In these cases, we manually rearrange the statements to ensure the query being used corresponds to the question. In a few cases, there is only one statement in the problem. In these cases, the context is empty.

AIME includes a latex representation of figures. However, these latex figures can leak information about the answer: for example, these latex figures can contain exact information about the lengths of the sides in a geometry problem, giving away the answer. In these cases we first ensure that the problem is solvable without the figure and then manually strip the figure latex from the problem context.

\section{Implementation Details}
\label{app:implementation_details}

We implement \name{} via function calling. When applying \name{}, the model is given access to two functions, \texttt{rethink\_memory} and  \texttt{finish\_rethinking}. The \texttt{rethink\_memory} function takes as input a new string, and replaces the current context $c$ and replaces the current context with the new string. The \texttt{finish\_rethinking} function terminates the \name process. The model is allowed to call the function \texttt{rethink\_memory} for up to 10 times.

\section{AIME main results by year}
\label{app:aime_main_by_year}

\begin{figure}[h!]
    \centering
    \includegraphics[width=\textwidth]{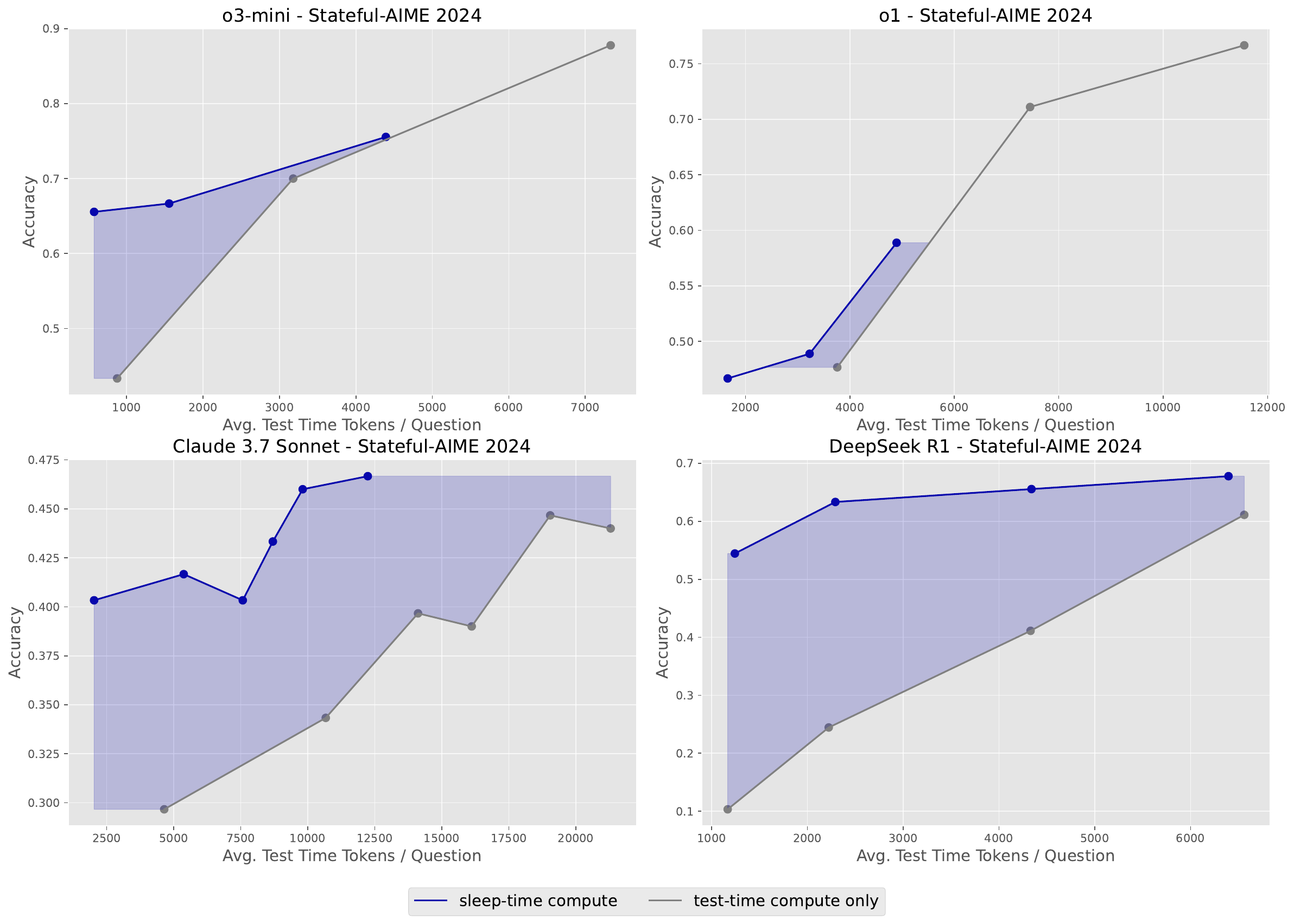}
    \caption{AIME 2024 main result}
    \label{fig:aime-main-result-2024}
\end{figure}

\begin{figure}[h!]
    \centering
    \includegraphics[width=\textwidth]{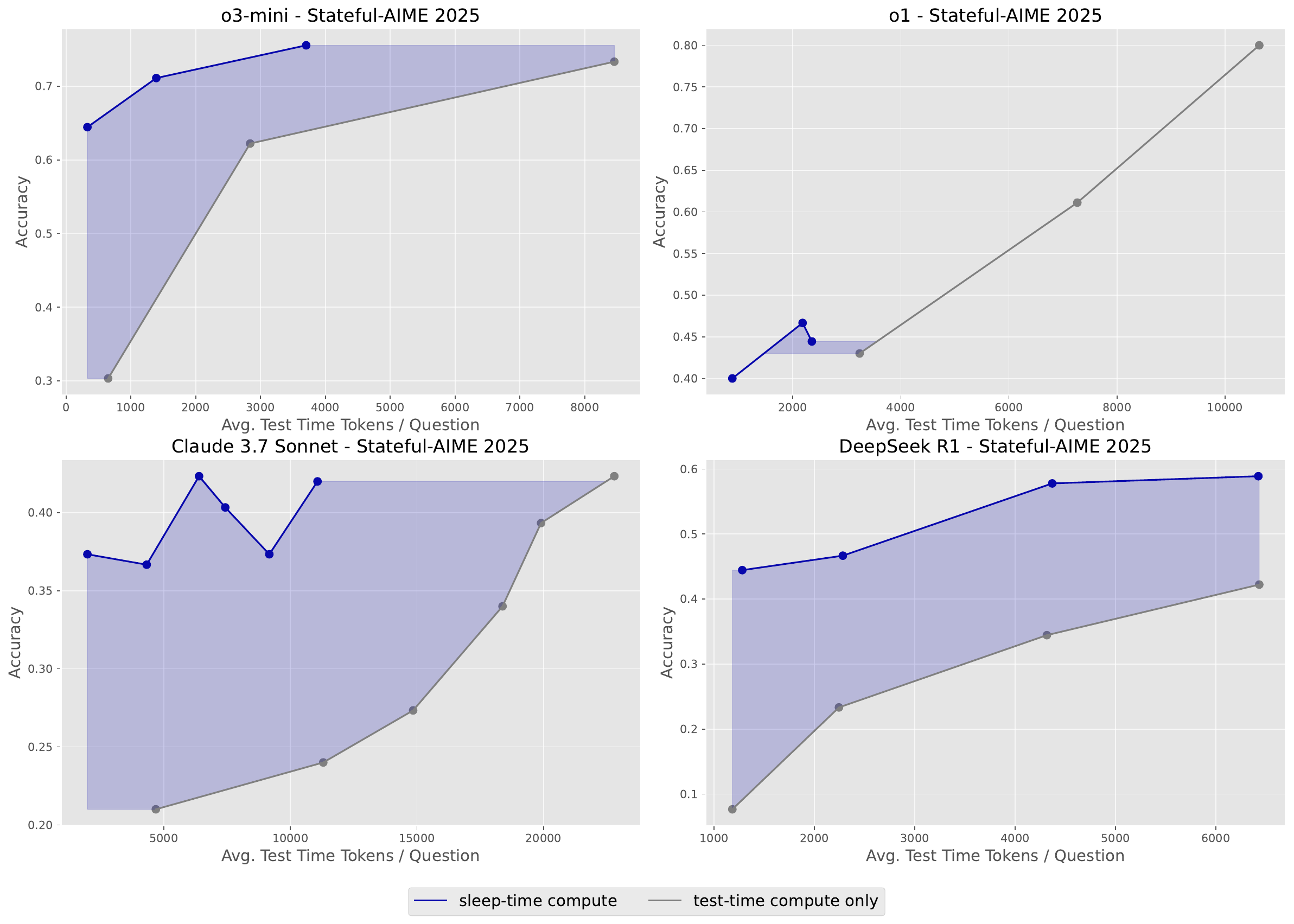}
    \caption{AIME 2025 main result}
    \label{fig:aime-main-result-2025}
\end{figure}

\section{AIME \name{} scaling results by year}
\label{app:aime_scaling_by_year}

\begin{figure}[h!]
    \centering
    \includegraphics[width=\textwidth]{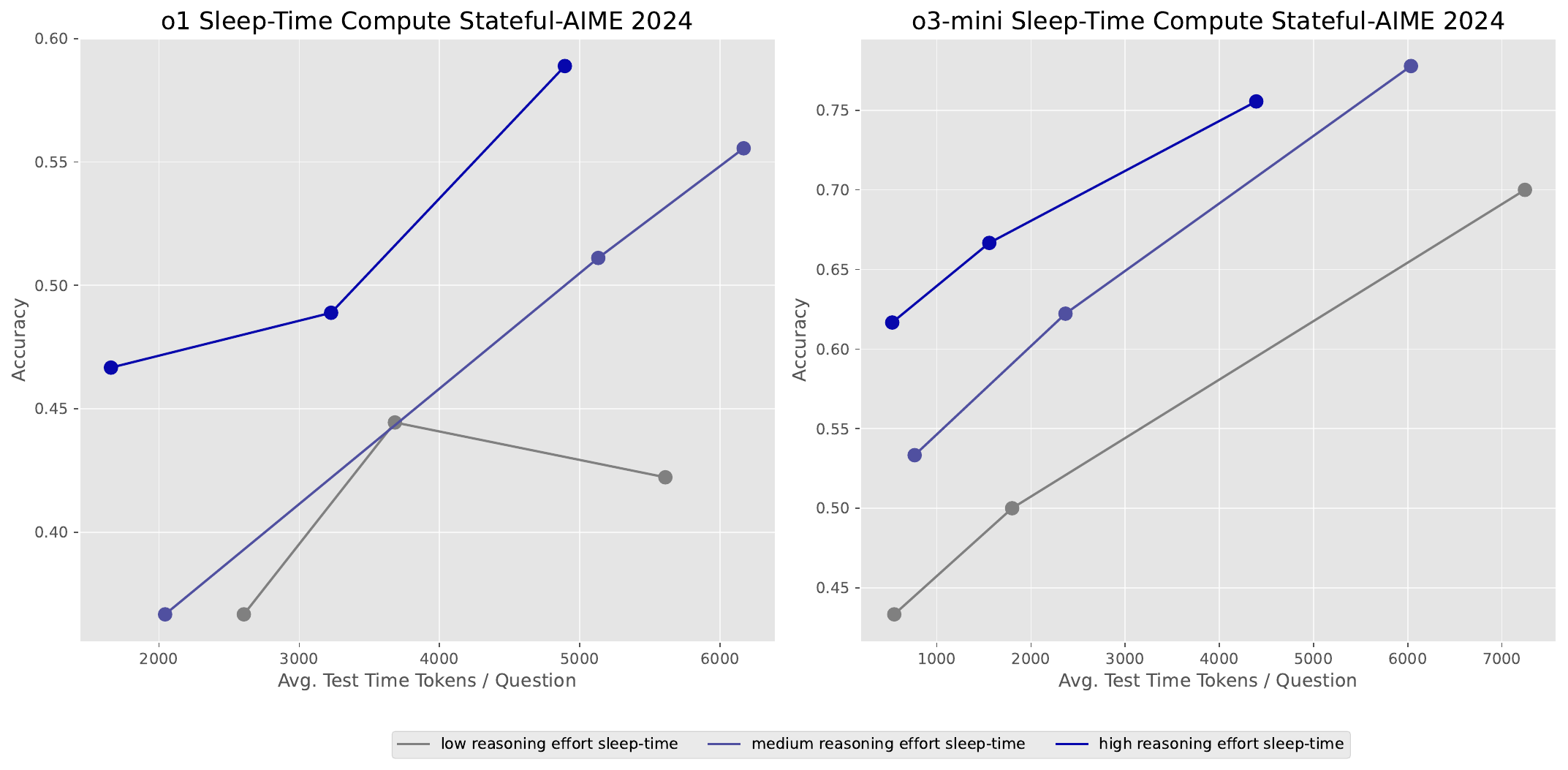}
    \caption{Scaling \name{} for \aimedataset{}2024.}
    \vspace{-0.2cm}
    \label{fig:aime_async_scaling}
\end{figure}

\begin{figure}[h!]
    \centering
    \includegraphics[width=\textwidth]{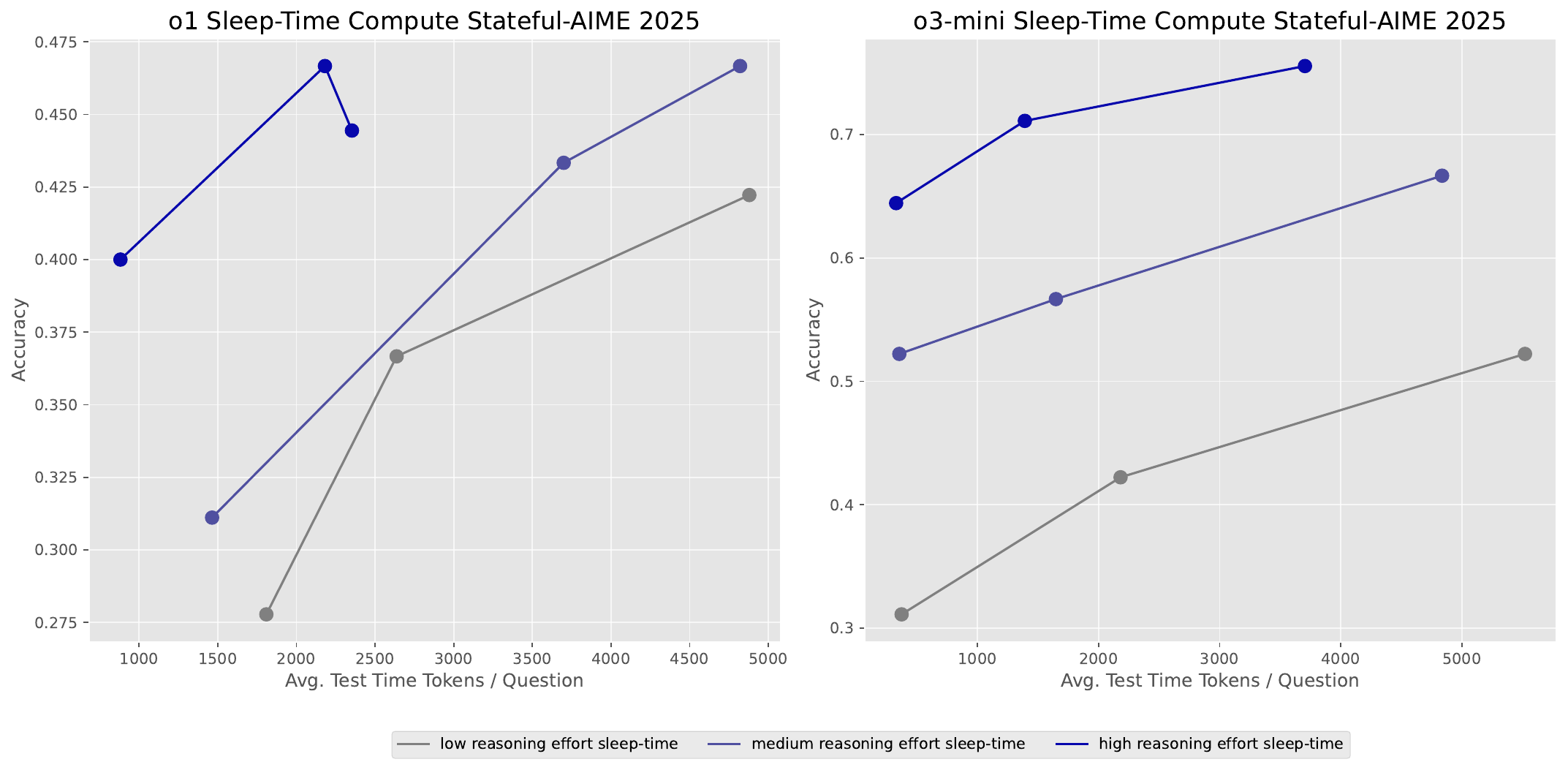}
    \caption{Scaling \name{}  on \aimedataset{}2025}
    \label{fig:aime_async_scaling}
\end{figure}

\end{document}